\documentclass{article}

\usepackage[preprint]{neurips_2025}

\usepackage[utf8]{inputenc} 
\usepackage[T1]{fontenc}    
\usepackage{hyperref}       
\usepackage{url}            
\usepackage{booktabs}       
\usepackage{amsfonts}       
\usepackage{nicefrac}       
\usepackage{microtype}      
\usepackage{xcolor}         
\usepackage{float}
\usepackage{newfloat}
\usepackage{multirow}
\usepackage{makecell}
\usepackage{colortbl}
\usepackage{dsfont}
\usepackage{amsmath}
\usepackage{enumitem}
\usepackage{algpseudocode} 
\usepackage[ruled,lined]{algorithm2e}
\usepackage{tcolorbox}       
\usepackage[normalem]{ulem}  
\useunder{\uline}{\ul}{}     
\usepackage{bbding}          
\usepackage{stackrel}        
\usepackage{todonotes}       
\usepackage{wrapfig}         
\usepackage{subcaption}      
\usepackage{bbm}             

\usepackage{listings}

\definecolor{codegray}{gray}{0.95}
\definecolor{commentgreen}{rgb}{0,0.6,0}
\definecolor{keywordblue}{rgb}{0.2,0.2,0.7}

\lstdefinestyle{pythonstyle}{
    language=Python,
    backgroundcolor=\color{codegray},
    basicstyle=\ttfamily\footnotesize,
    keywordstyle=\color{keywordblue}\bfseries,
    commentstyle=\color{commentgreen}\itshape,
    stringstyle=\color{purple},
    showstringspaces=false,
    breaklines=true,
    tabsize=4,
    captionpos=b
}

\newcommand{\inlineeqnum}[1]{%
  \refstepcounter{equation}  
  \label{#1}                 
  \textnormal{(\theequation)}
}

\renewcommand{\todo}[1]{\iffalse #1 \fi{\color{blue} \textbf{[TODO]}}}

\newcommand{\leftrarrows}{\mathrel{\raise.9ex\hbox{\oalign{%
  $\scriptstyle\leftarrow$\cr
  \vrule width0pt height.5ex$\hfil\scriptstyle\relbar$\cr}}}}
\newcommand{\lrightarrows}{\mathrel{\raise.9ex\hbox{\oalign{%
  $\scriptstyle\relbar$\hfil\cr
  $\scriptstyle\vrule width0pt height.5ex\smash\rightarrow$\cr}}}}
\newcommand{\Rrelbar}{\mathrel{\raise.9ex\hbox{\oalign{%
  $\scriptstyle\relbar$\cr
  \vrule width0pt height.5ex$\scriptstyle\relbar$}}}}

\newtheorem{theorem}{Theorem}

\usepackage{tikz}

\newcommand{\method}{PPAD}

\DeclareMathOperator*{\argminA}{arg\,min}

\title{\resizebox{\textwidth}{!}{Multimodal LLM-Guided Semantic Correction in Text-to-Image Diffusion}}

\author{
Zheqi Lv\textsuperscript{1} \quad
Junhao Chen\textsuperscript{1} \quad
Qi Tian\textsuperscript{1} \quad
Keting Yin\textsuperscript{1} \quad
Shengyu Zhang\textsuperscript{1} \quad
Fei Wu\textsuperscript{1} \\
\textsuperscript{1}Zhejiang University \\
\texttt{\{zheqilv, chenjunhao100, tianqics, yinkt, sy\_zhang, wufei\}@zju.edu.cn}\\ 
\textcolor{red}{\url{https://github.com/HelloZicky/PPAD}}
}

\begin{document}

\maketitle

\begin{abstract}
\label{sec:abstract}
Diffusion models have become the mainstream architecture for text-to-image generation, achieving remarkable progress in visual quality and prompt controllability. However, current inference pipelines generally lack interpretable semantic supervision and correction mechanisms throughout the denoising process. Most existing approaches rely solely on post-hoc scoring of the final image, prompt filtering, or heuristic resampling strategies—making them ineffective in providing actionable guidance for correcting the generative trajectory. As a result, models often suffer from object confusion, spatial errors, inaccurate counts, and missing semantic elements, severely compromising prompt-image alignment and image quality.
To tackle these challenges, we propose \textit{MLLM Semantic-Corrected Ping-Pong-Ahead Diffusion} (\textit{PPAD}), a novel framework that, for the first time, introduces a Multimodal Large Language Model (MLLM) as a semantic observer during inference. PPAD performs real-time analysis on intermediate generations, identifies latent semantic inconsistencies, and translates feedback into controllable signals that actively guide the remaining denoising steps. The framework supports both inference-only and training-enhanced settings, and performs semantic correction at only extremely few diffusion steps, offering strong generality and scalability. Extensive experiments  demonstrate PPAD's significant improvements.
\end{abstract}
\section{Introduction}
\label{sec:introduction}

\begin{figure}[!th]
    \centering
    \includegraphics[width=\textwidth]{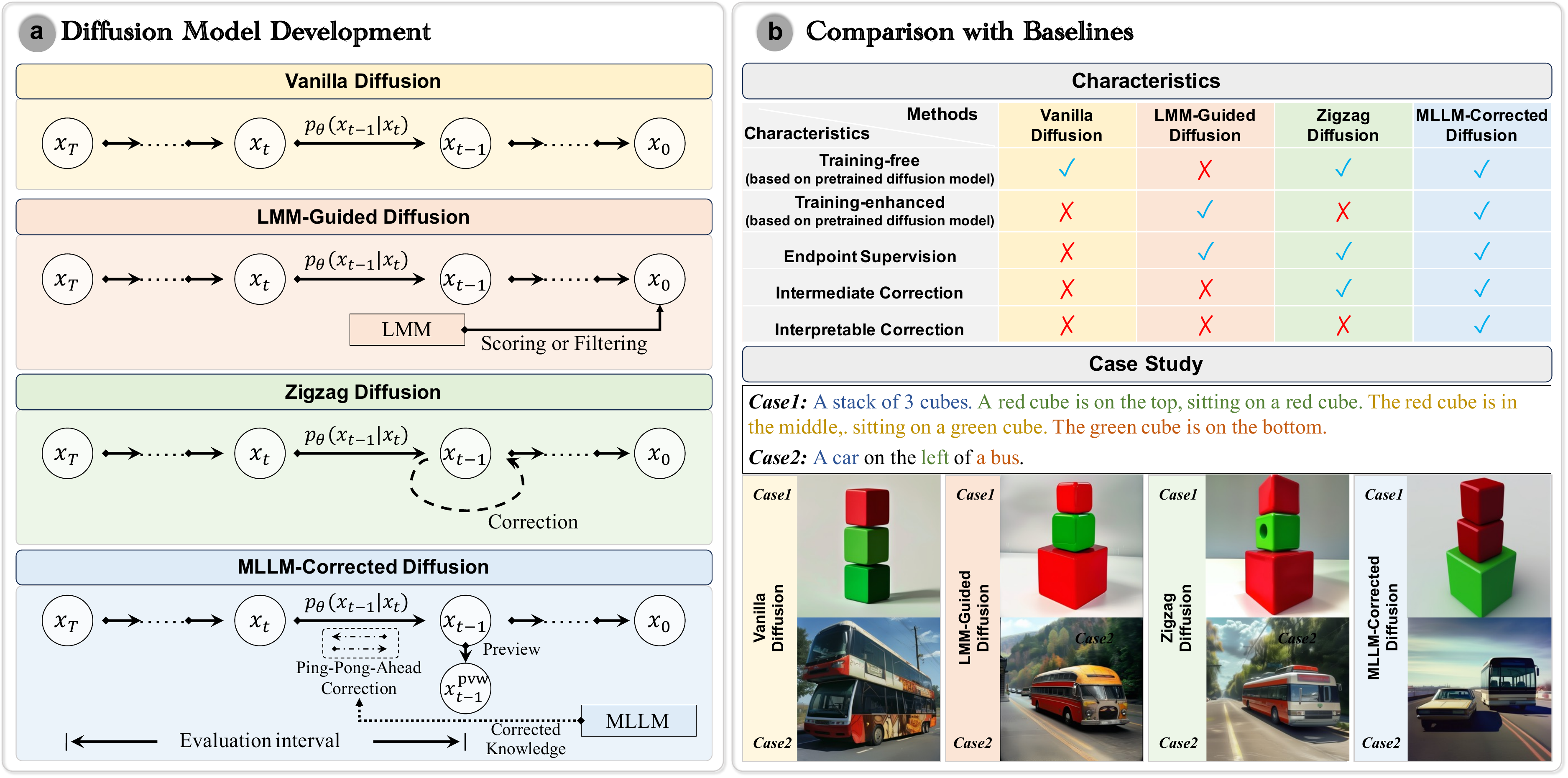}
    \vspace{-0.55cm}
    \caption{Background and brief comparison of the baselines and our \method{}. (a) compares the workflows of four diffusion methods, while (b) summarizes their characteristics and presents generated images under the same prompt, demonstrating that our method effectively corrects semantic errors and guides generation in the right direction.}
    \label{fig:introduction}
    \vspace{-0.65cm}
\end{figure}

Diffusion models have become the cornerstone of text-to-image generation, achieving unprecedented visual quality and controllability~\cite{rombach2022high, podell2023sdxl, li2024hunyuan}. However, as illustrated in Figure~\ref{fig:introduction}, current inference pipelines remain vulnerable to semantic misalignment—failing to correct object confusion, spatial errors, and missing entities during the denoising process.

As shown in Figure~\ref{fig:introduction}(a) and (b), vanilla diffusion models \cite{ho2020denoising} lack interpretive feedback and corrective signals during the generation process, often resulting in inadequate semantic understanding. LMM-guided diffusion methods \cite{nichol2021glide, saharia2022photorealistic, black2023diffusiondpo} incorporate CLIP-based scorers or LLM-generated scene hints, but they only function before or after generation, lacking any inspection or correction during inference. Zigzag Diffusion~\cite{huang2023reflected, bai2024zigzag} introduces alternating forward and reverse denoising steps to refine latent representations, but its corrections are implicit, lack interpretability, and rely on heuristic resampling, which makes the refinement direction unstable.
As illustrated in Figure~\ref{fig:introduction}(b), the two cases involve multiple semantic attributes such as color, position, and shape, yet existing methods struggle to accurately map these semantics into the generated image. (More cases are provided in Figure~\ref{fig:generation_case1}.)

To address these shortcomings, we propose MLLM Semantic Corrected Ping-Pong-Ahead Diffusion (PPAD), which leverages a Multimodal Large Language Model (MLLM) as an intelligent ``observer'' and ``corrector'' in the diffusion process. Our method performs on-the-fly semantic analysis of partial generations and injects high-level feedback to adjust the sampling trajectory. This yields an efficient guidance mechanism that not only keeps the generation on track with the input semantics, but also makes the correction process transparent and controllable. 
PPAD consists of four modules: (1) \emph{Lookahead Sketch Generator}, which generates a preview sketch of the upcoming image to support early-stage semantic feedback; (2) \emph{Semantic Consistency Checker}, which evaluates the semantic alignment between the current output and the original prompt; (3) \emph{Corrective Prompt Synthesizer}, which produces a refined prompt and highlights missing semantic elements; (4) \emph{Ping-Pong-Ahead}, which injects Semantic-Corrected Knowledge (SCK) into the diffusion process through rollback and resampling.
PPAD also supports optional training strategies including supervised fine-tuning (PPAD-SFT) and preference optimization (PPAD-DPO), further aligning image features with semantic corrections. The Lookahead Sketch Generator is designed to prepare for the generation of Semantic Corrected Knowledge (SCK), which is produced by the Semantic Consistency Checker and Corrective Prompt Synthesizer. The Ping-Pong-Ahead is responsible for injecting and utilizing the SCK to correct the denoising trajectory. \textit{\uline{Note that PPAD supports both inference-only and training-enhanced settings, and performs semantic correction at only extremely few diffusion steps, offering strong generality and scalability.}}

Our contributions are summarized as follows:
\begin{itemize}
    \item We pioneer the use of MLLMs during the diffusion process, enabling interpretable and progressive correction that significantly enhances both semantic alignment and visual quality.
    \item We design the Lookahead Sketch Generator and Ping-Pong-Ahead mechanisms to enable the extraction and injection of SCK during the diffusion process.
    \item We provide a unified framework that supports inference-only and training-enhanced settings, adaptable to mainstream diffusion models.
    \item Extensive experiments across various prompt types, evaluation metrics, and diffusion backbones demonstrate that PPAD achieves significant improvements over existing methods.
\end{itemize}

\section{Related Work}
\label{sec:related_work}

Modern MLLMs adopt a Perceiver-to-LLM design: ViT features \cite{dosovitskiy2020image} are mapped via learnable queries (BLIP-2 \cite{li2023blip}, Qwen-VL \cite{QWen-VL}) or linear connectors (LLaVA \cite{liu2023improved,liu2024llavanext}) into frozen LLMs. Alignment is improved with Flamingo’s Resampler+XATTN \cite{alayrac2022flamingo} and InternLM’s Partial-LoRA \cite{zhang2023internlm}. Closed models like GPT-4V \cite{gpt4v}, Gemini-1.5 \cite{gemini1.5}, Claude-3V \cite{Claude3v}, and MM1 \cite{mm1} lead performance; open ones (VisionLLM \cite{visionllm}, CogVLM \cite{cogvlm}, LLaMA-Adapter V2 \cite{llamaadapterv2}, ShareGPT-4V \cite{sharegpt4v}, LLaVA variants \cite{llava,llavanext}) are catching up.
MLLMs now handle video, audio, and point clouds \cite{sun2024video,lyu2023macaw,wu2023next}, with domain-specific versions for medicine and documents \cite{li2023llavamed,lan2025gem,ye2023mplugdoc}. Instruction tuning (e.g., InstructBLIP \cite{dai2023instructblip}) enables strong performance with minimal updates.
Yet few works exploit MLLMs for in-process semantic correction. We instead employ them as inspectors that analyze intermediate outputs and inject interpretable, directional feedback during denoising.

\noindent \textbf{Diffusion Models.}
DDPMs learn to reverse Gaussian noise \cite{sohl2015deep,ho2020denoising}. A score-based SDE view unifies discrete and continuous-time sampling \cite{song2019generative,song2020score}, while objective improvements enhance output fidelity \cite{nichol2021improved,kingma2021vdms}. Diffusion models now outperform GANs on FID \cite{dhariwal2021diffusion}.
Efficiency has improved via DDIMs \cite{song2020denoising}, FastDPM \cite{kong_fast_2021}, and one-step distillation \cite{salimans_progressive_2022}; LDMs operate in latent space \cite{rombach2022high}, and cascaded models scale to $1024^2$ \cite{podell2023sdxl,saharia2022imagen}. Applications span histopathology, satellite imagery, and gigapixel generation \cite{moghadam2023morphology,xu2023vit,muller2023multimodal,yellapragada2023pathldm,espinosa2023generate,sebaq2023rsdiff,aversa2023diffinfinite}.
\textit{Multimodal-encoder-guided diffusion} injects static prompt embeddings from frozen encoders (e.g., CLIP, ALIGN, BLIP) into the UNet, as in GLIDE, LDM, and ControlNet \cite{nichol2021improved,rombach2022high,zhang2023adding}, but lacks step-wise correction.
\textit{LLM-conditioned diffusion} (e.g., DALL-E 2/3, Imagen) encodes prompts with large LLMs but retains fixed embeddings throughout \cite{ramesh2021dalle,saharia2022imagen}, even when using classifier-free guidance or distillation \cite{ho2022classifier,salimans_progressive_2022}.
\textit{MLLMs-guided diffusion} (BLIP-2, Qwen-VL, LLaVA) couples vision encoders with frozen LLMs for post-hoc reasoning \cite{li2023blip,qwenvl,liu2023improved}, typically as critics (ImageReward, PickScore, Diffusion-DPO) or for prompt rewriting (RPG, EasyGen) \cite{xu2023imagereward,kirstain2023pickscore,black2023diffusiondpo,liu2024rpg}.
However, most diffusion systems encode prompt semantics statically and apply reranking only after sampling—insufficient to correct denoising drift. In contrast, our approach previews intermediate generations, diagnoses semantic gaps, and injects dynamic feedback to iteratively steer sampling.
\section{Methodology}
\label{sec:method}

\subsection{Problem Formulation and Notations}
\label{sec:problem}

\noindent\textbf{Data.} 
$\mathcal{D}_{\text{train}}=\{(\mathbf{p}^{(i)}, \mathbf{x}^{(i)})\}_{i=1}^{N}$ stores prompt–image pairs for Supervised Fine-Tuning (SFT) and Direct Preference Optimization (DPO), while $\mathcal{D}_{\text{test}}=\{\mathbf{p}^{(j)}\}_{j=1}^{M}$ provides prompts only; at test time we synthesize an image $\hat{\mathbf{x}}^{(j)}$ for each $\mathbf{p}^{(j)}$. When discussing DPO, two images are involved, where the superscripts ``$+$'' and ``$-$'' denote the positive and negative samples, respectively.

\noindent\textbf{Model.} We define $\mathtt{E}$ as the text encoder. The diffusion backbone $\mathcal{M}^{\textsc{dm}}_{\theta}(\mathbf{z}, t, \mathbf{c})$, parameterized by $\theta$, denoises latent variable $\mathbf{z}$ at timestep $t$ under condition $\mathbf{c}$. The LMM $\mathcal{M}^{\textsc{lmm}}$ outputs a scalar alignment logit for training supervision. The MLLM $\mathcal{M}^{\textsc{mllm}}$ produces both a scalar score and token-level critique for real-time inference correction.

\noindent\textbf{Feature and Loss.} 
We denote the latent at step $t$ as $\mathbf{x}_{t}$, with variance schedule $\beta_{t}$ controlling forward noise and reverse variance $\sigma^2_{t}$ governing the stochasticity of denoising. Gaussian noise is denoted by $\boldsymbol{\epsilon}$. The LMM-based training score is defined as $\mathcal{R}^{\textsc{lmm}} = \mathcal{M}^{\textsc{lmm}}(\hat{\mathbf{x}}_{0}, \mathbf{p})$, based on predicted output $\hat{\mathbf{x}}_{0}$ and prompt $\mathbf{p}$. Zigzag diffusion introduces a self-reflection score $\mathcal{R}^{\textsc{zz}}_t = \mathcal{M}^{\textsc{zz}}(\mathbf{x}_{t}, \tilde{\mathbf{x}}_{t})$, where $\tilde{\mathbf{x}}_{t}$ is obtained by reversing one step and resampling from $\mathbf{x}_{t}$. The semantic correction signal from MLLM is $\mathcal{R}^{\textsc{mllm}}_t = \mathcal{M}^{\textsc{mllm}}(\mathbf{x}_{t}, \mathbf{p})$, providing interpretable guidance during inference.

\noindent\textbf{Formula.} 
To clarify the distinctions between our method PPAD and other diffusion models such as Vanilla Diffusion (VD), LMM-Guided Diffusion (LGD), and Zigzag Diffusion (ZZD), we provide concise formulaic comparisons in the following.
\begin{flushleft}
\textit{VD}
\hspace{3.05em}
\hspace{-0.5em}
\textbf{Train:}\;
$\displaystyle
  \mathcal{L}_{\textsc{sft}}(\mathbf{x}_{0}, \hat{\mathbf{x}}_{0})
$\hspace{5.6em}\inlineeqnum{eq:dm_train}
\hspace{2.4em}
\textbf{Inference:}\;
$\displaystyle
  \mathbf{x}_{t-1}=%
  \mathcal{M}^{\textsc{dm}}_{\theta}(\mathbf{x}_{t},\mathbf{p})
$\hspace{2.4em}\inlineeqnum{eq:dm_inf}

\textit{LGD}
\hspace{2.6em}
\hspace{-0.7em}
\textbf{Train:}\;
$\displaystyle
  \mathcal{L}_{\textsc{dpo}}(\hat{\mathbf{x}}_{0}^{+}, \hat{\mathbf{x}}_{0}^{-},
  \mathcal{R}^{\textsc{lmm}})
$\hspace{2.2em}\inlineeqnum{eq:lmm_train}
\hspace{2.4em}
\textbf{Inference:}\;
$\displaystyle
  \mathbf{x}_{t-1}=%
  \mathcal{M}^{\textsc{dm}}_{\theta}(\mathbf{x}_{t},\mathbf{p})
$\hspace{2.4em}\inlineeqnum{eq:lmm_inf}

\textit{ZZD}
\hspace{2.5em}
\hspace{-0.5em}
\textbf{Inference:}\;
$\displaystyle
  \mathbf{x}_{t-1}=%
\mathcal{M}^{\textsc{dm}}_{\theta}\!\bigl(\mathbf{x}_{t},
  \mathcal{R}^{\textsc{zz}},\mathbf{p}\bigr)
$\hspace{18.5em}\inlineeqnum{eq:zz_inf}

\textit{PPAD}
\hspace{1.95em}
\hspace{-0.5em}
\textbf{Train:}\;
$\displaystyle
  \mathcal{L}_{\textsc{sft}}(\mathbf{x}_{0}, \hat{\mathbf{x}}_{0},
  \mathcal{R}^\textsc{mllm})
  +\mathcal{L}_{\textsc{dpo}}(\hat{\mathbf{x}}_{0}^{+},
  \hat{\mathbf{x}}_{0}^{-}, \mathcal{R}^{\textsc{mllm}})
$\hspace{12.2em}\inlineeqnum{eq:ppad_train}

\hspace{4.6em}
\hspace{-0.5em}
\textbf{Inference:}\;
$\displaystyle
  \mathbf{x}_{t-1}=%
  \mathcal{M}^{\textsc{dm}}_{\theta}\!\bigl(\mathbf{x}_{t},
  \mathcal{R}^{\textsc{mllm}},\mathbf{p}\bigr)
$\hspace{17.4em}\inlineeqnum{eq:ppad_inf}
\end{flushleft}
\noindent\emph{Read-off:}
Eq.~\eqref{eq:dm_train} and \eqref{eq:dm_inf} represent the standard diffusion model.
Eq.~\eqref{eq:lmm_train} and \eqref{eq:lmm_inf} provide only endpoint supervision during training, with no guidance during inference or the denoising process.
Eq.~\eqref{eq:zz_inf} attempts to provide correction during inference, but relies on differences from repeated denoising samples, making it unstable and hard to interpret.
Eq.~\eqref{eq:ppad_inf} enables interpretable semantic correction under the inference-only setting, combining both process-level guidance and endpoint supervision.
Eq.~\eqref{eq:ppad_train} indicates that PPAD is also compatible with training-enhanced settings.

\subsection{Preliminary}
\label{sec:prelim}

\textit{Forward corruption.}
Let $\beta_{t}\in(0,1)$ be a variance schedule
($1\!\le\!t\!\le\!T$).  
Define $\alpha_{t}=1-\beta_{t}$,\; $\bar\alpha_{t}=\prod_{s=1}^{t}\alpha_{s}$,\; $\sigma_{t}^{2}=\beta_{t}\,(1-\bar\alpha_{t-1})/(1-\bar\alpha_{t})$ with $\bar\alpha_{0}=1$.
For a clean image $\mathbf{x}_{0}\!\in\!\mathbb{R}^{H\times W\times3}$ the forward (noising) Markov chain is
\begin{equation}
q(\mathbf{x}_{t}\mid\mathbf{x}_{t-1})
=\mathcal{N}\!\bigl(
      \sqrt{\alpha_{t}}\mathbf{x}_{t-1},
      \beta_{t}\mathbf{I}\bigr),
\qquad
t=1,\dots,T.
\label{eq:ddpm_forward}
\end{equation}

\textit{Training Procedure.}
Sampling $(\mathbf{x}_{0},\mathbf{p})\!\sim\!\mathcal{D}_{\text{train}}$ and a timestep $t\!\sim\!\mathrm{Unif}\{1,\dots,T\}$, we form
\begin{equation}
\mathbf{x}_{t}=\sqrt{\alpha_{t}}\mathbf{x}_{t-1}+\sqrt{1-\alpha_{t}}\;\boldsymbol{\epsilon}_{t},\;\boldsymbol{\epsilon}_{t}\sim\mathcal{N}(\mathbf{0},\mathbf{I})
\label{eq:ddpm_one_step_noise}
\end{equation}

We define the above transformation as:
$
\mathcal{F}_{t-1\!\to t}(\mathbf{x}_{t-1},t).
$

By stacking this equation from timestep $1$ to $t$, we obtain the standard form:
$\mathbf{x}_{t}
=\sqrt{\bar\alpha_{t}}\mathbf{x}_{0}
 +\sqrt{1-\bar\alpha_{t}}\;\boldsymbol{\epsilon}$.
Then minimise the DM loss to get $\theta$
\begin{equation}
\argminA_\theta \mathcal{L}_{\textsc{sft}}
=\argminA_\theta \Bigl\|
\boldsymbol{\epsilon}-%
\mathcal{M}^{\textsc{dm}}_{\theta}
      (\mathbf{x}_{t},t,\mathtt{E}(\mathbf{p}))\Bigr\|^{2}.
\label{eq:ddpm_sft}
\end{equation}

\textit{Inference Procedure.}
Given a prompt $\mathbf{p}$, the text encoder provides
$\mathbf{c}=\mathtt{E}(\mathbf{p})$.  
We sample $\mathbf{x}_{T}\sim\mathcal{N}(\mathbf{0},\mathbf{I})$ and can get $\mathbf{x}_{t-1}$ based on $\mathbf{x}_{t}$ as follows,
\begin{equation}
\mathbf{x}_{t-1}=
\sqrt{\bar\alpha_{t-1}} \cdot \left( \frac{\mathbf{x}_t - \sqrt{1 - \bar\alpha_t} \cdot \mathcal{M}^{\textsc{dm}}_{\theta}(\mathbf{x}_t, t, \mathbf{c}) }{ \sqrt{\bar\alpha_t} } \right)
+ \sqrt{1 - \bar\alpha_{t-1}} \cdot \mathcal{M}^{\textsc{dm}}_{\theta}(\mathbf{x}_t, t, \mathbf{c})
\label{eq:ddpm_one_step_denoise}
\end{equation}

\subsection{MLLM Semantic-Corrected Ping-Pong-Ahead Diffusion}
\label{sec:ppad}

\begin{figure}[!t]
    \centering
    \includegraphics[width=\textwidth]{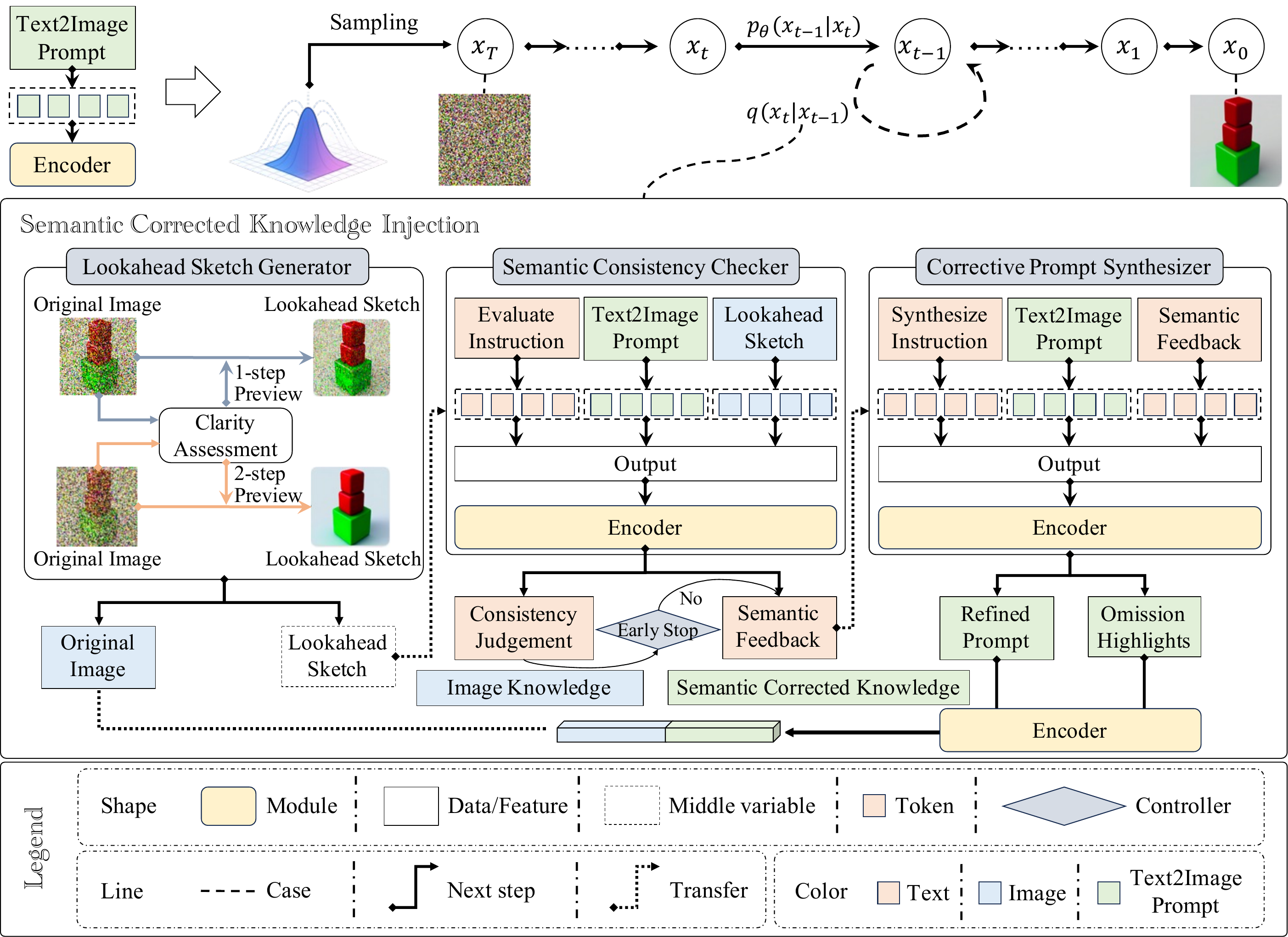}
    \vspace{-0.5cm}
    \caption{Overview of the proposed method.}
    \label{fig:method}
    \vspace{-0.6cm}
\end{figure}

As shown in Figure~\ref{fig:method}, \method{} consists of four components: Lookahead Sketch Generator, Semantic Consistency Checker, Corrective Prompt Synthesizer, and Ping-Pong-Ahead. We introduce them and the optional training procedure in the following, enabling real-time semantic correction during both denoising and output stages via MLLM feedback. It supports a training-free, plug-and-play deployment by integrating $\mathcal{M}^{\textsc{mllm}}$ into the diffusion backbone. For further alignment between semantic feedback and latent features, \method{} also supports post-training with standard methods like SFT and DPO.

\subsubsection{Lookahead Sketch Generator}\label{Lookahead Sketch Generator}
We introduce the \textit{Lookahead Sketch Generator} (LSG), which addresses the challenge that ambiguous intermediate states during the diffusion process hinder the MLLM from producing reliable semantic feedback. As shown in the second row of Table~\ref{tab:ablation_module} in Sec.~\ref{sec:ablation_study}, the absence of LSG leads to misleading or unstable semantic correction.

At each denoising step $t$, given the latent variable $\mathbf{x}_{t}$, we determine the lookahead step count $k(\mathbf{x}_{t}) \in \{1, 2\}$ based on the signal-to-noise ratio (SNR). We then perform a constrained denoising rollout of $k(\mathbf{x}_{t})$ steps, yielding a \textit{Lookahead Sketch}—a rough approximation of the future generation trajectory. Although this sketch is not perfectly accurate, it captures sufficient semantic structure to support downstream evaluation.
This sketch is then fed into the MLLM to enable early correction, even when $\mathbf{x}_{t}$ is too noisy. The threshold $\gamma$ controls whether to perform lookahead. The full process of LSG is formally defined as:
\begin{equation}
\label{eq:preview_operator}
\mathbf{x}^{\text{pvw}}_{t}
=
\underbrace{\mathcal{M}^{\textsc{dm}}_{\theta} (\mathbf{x}_{k(t)},k(t),\,\mathtt{E}(\mathbf{p}))}_{\text{Lookahaed Sketch Generation}},
\qquad
k(t)=2-\mathbf{1}[\mathrm{SNR}(t)>\gamma].
\end{equation}
\subsubsection{Semantic Corrected Knowledge}\label{Semantic Corrected Knowledge}
We extract the Semantic Corrected Knowledge (SCK) through two steps: Semantic Consistency Checker (SCC) and Corrective Prompt Synthesizer (CPS).

\noindent \textit{Semantic Consistency Checker.}
We define a correction interval from $t_e$ to $t_s$, during which $\mathcal{M}^{\textsc{mllm}}$ is queried every $\Delta$ steps to assess the current generation trajectory. At each step $t$, we evaluate the \textit{Lookahead Sketch} $\mathbf{x}^{\text{pvw}}_{t}$ produced by the Lookahead Sketch Generator. The SCC outputs a tuple $\mathcal{R}^{\textsc{scc}}_t = \{s_t, S_t\}$, where $s_t$ is a binary consistency score indicating whether the current trajectory aligns with the text prompt $\mathbf{p}$, and $S_t$ is the semantic feedback detailing aspects such as color, shape, count, and spatial structure.
If $s_t$ exceeds the threshold $\tau_{\text{stop}}$, we regard the generation as semantically aligned and skip all remaining SCC queries in this interval, continuing with standard denoising as in Eq.~\ref{eq:dm_inf}. Otherwise, $S_t$ is passed to the Corrective Prompt Synthesizer for further refinement. This process can be formally defined as:
\begin{equation}
\mathcal{R}^\textsc{mllm}_t = \{s_{t}, S_{t}\}
=\mathcal{M}^{\textsc{mllm}}\!\left(\mathbf{x}^{\text{pvw}}_{t},\mathbf{p}\right),
\quad
\text{if }s_{t}\ge\tau_{\text{stop}}\;\Longrightarrow\;
\text{early stopping}.
\label{eq:evaluate}
\end{equation}
The values of $t$ satisfy $t \in \{\, t_s - k \cdot \Delta \;\mid\; k = 0, 1, 2, \dots, \left\lfloor \tfrac{t_s - t_e}{\Delta} \right\rfloor \,\}$. $\tau_{st   op}$ is the consistency score threshold used to determine early stopping.

\noindent \textit{Corrective Prompt Synthesizer.}
If the early stopping condition is not met in the previous stage, the semantic feedback $S_t$ is passed to this module and rewritten by $\mathcal{M}^{\textsc{mllm}}$ into two components: a \textit{Refined Prompt} $s^+_t$, which reinforces missing or underspecified attributes such as style, color, quantity, or shape; and an \textit{Omission Highlights} list $s^-_t$, which explicitly enumerates unmet elements implied by the original prompt.
These two textual outputs are then encoded using a text encoder into conditioning vectors $\mathbf{c}^+_t$ and $\mathbf{c}^-_t$, respectively. The resulting embeddings are concatenated or composed to guide the backbone diffusion model $\mathcal{M}^{\textsc{dm}}_{\theta}$ in subsequent denoising steps.


\subsubsection{Ping-Pong-Ahead}\label{Ping-Pong-Ahead}

After obtaining the SCK, the primary challenge becomes how to effectively inject it. To address this, we design Ping-Pong-Ahead (PPA) to accomplish this step.
At first, we introduce the reverse and forward operators. 
For the forward operator and the reverse operator, we adopt Eq.~\ref{eq:ddpm_one_step_noise} and Eq.~\ref{eq:ddpm_one_step_denoise}, respectively.
Given the current latent variable $\mathbf{x}_{t-1}$, we perform:
\begin{subequations}
\label{eq:ppa_ddpm}
\begin{align}
\textbf{Ping} \text{ (\textit{back-step})}:\;&
\widetilde{\mathbf{x}}_{t}
       =\widetilde{\mathcal{F}}_{t-1\to t}(\mathbf{x}_{t-1}),
       \label{eq:ping_ddim}\\
\textbf{Pong} \text{ (\textit{forward-step})}:\;&
\widetilde{\textbf{c}}_{t}=\mathtt{E}(s^{+}_{t})\;\Vert\;\mathtt{E}(s^{-}_{t}),
\quad
\widetilde{\mathbf{x}}_{t-1} =\mathcal{F}_{t\to t-1}\!(\widetilde{\mathbf{x}}_{t},\widetilde{\textbf{c}}_{t}),
\label{eq:pong_ddpm}\\
\textbf{Ahead} \text{ (\textit{forward-step})}:\;&
\mathbf{x}_{t-2}=\mathcal{F}_{t-1\to t-2}
(\widetilde{\mathbf{x}}_{t-1},\mathbf{c}),
\label{eq:ahead}
\end{align}
\end{subequations}
where \(s^{+}_{t}\) (refined prompt) and \(s^{-}_{t}\) (omission highlights) come from the SCC–CPS pipeline, and “\(\Vert\)” is concatenation.  
Eq.\,\eqref{eq:ping_ddim} is deterministic; \eqref{eq:pong_ddpm} introduces fresh noise via $\sigma_{t}$; \eqref{eq:ahead} resumes standard DM with the original conditioning $\mathbf{c}$. Same as the above, the values of $t$ satisfy $t \in \{\, t_s + k \cdot \Delta \;\mid\; k = 0, 1, 2, \dots, \left\lfloor \tfrac{t_e - t_s}{\Delta} \right\rfloor \,\}$. At timesteps not included in this set, $\mathbf{x}_t$ is denoised using the standard DM procedure (Eq.~\ref{eq:ddpm_one_step_denoise}). Through the PPA mechanism, we can effectively inject the SCK obtained at the current latent variable (e.g., $\mathbf{x}_{t-1}$) into the preceding latent variable $\mathbf{x}_t$, thereby correcting the denoising process from $\mathbf{x}*t \rightarrow \mathbf{x}_{t-1}$.
\textit{Note that it is not necessary to perform SCK extraction and PPA at every timestep; applying them once every $\Delta$ steps within the interval $t_s \sim t_e$ is sufficient to effectively guide and refine the denoising direction.}

\subsubsection{Training Procedure (Optional)}

\textit{\method{}-SFT.}
After one Ping–Pong cycle we decode
$\hat{\mathbf{x}}_{0}$ from $\widetilde{\mathbf{x}}_{t-1}$ using standard
steps and minimise
\begin{equation}
\mathcal{L}_{\textsc{ppad-sft}}
=\Bigl\|
\boldsymbol{\epsilon}_{t}-
\mathcal{M}^{\textsc{dm}}_{\theta}(\widetilde{\mathbf{x}}_{t},t,\widetilde{c}_{t})\Bigr\|^{2},
\label{eq:ppad_sft_new}
\end{equation}
which is the standard DM loss evaluated on the \emph{Ping} latent and with
the corrective embedding.

\textit{\method{}-DPO.}
Positive/negative images are both produced by \emph{full PPAD sampling}
(including Ping–Pong corrections).  Preference is judged by the same MLLM
score used inside the loop:
\begin{equation}
\mathcal{L}_{\textsc{ppad-dpo}}
=-\log
\frac{\exp S_{\psi}(\hat{\mathbf{x}}^{+}_{0},\mathbf{p})}{
      \exp S_{\psi}(\hat{\mathbf{x}}^{+}_{0},\mathbf{p})
     +\exp S_{\psi}(\hat{\mathbf{x}}^{-}_{0},\mathbf{p})},
\label{eq:dpo_final}
\end{equation}
$\hat{\mathbf{x}}^{+}_{0},\hat{\mathbf{x}}^{-}_{0}\;
\text{decoded from }(\widetilde{\mathbf{x}}_{t-1})^{+},(\widetilde{\mathbf{x}}_{t-1})^{-}
\text{ via standard DM steps.}$
Everything back-propagates through
$\mathcal{M}^{\textsc{mllm}}\!\rightarrow\!\mathcal{M}^{\textsc{dm}}_{\theta}$.
Zero-shot \method{} already improves over baselines; training further tightens
image–text alignment.

\subsection{Theoretical Foundations of \method{}}
In Section \ref{Lookahead Sketch Generator} and Section \ref{Semantic Corrected Knowledge}, we introduced a SNR-constrained denoising framework. Theorem \ref{theorem:1} formally establishes that this methodology bounds the cumulative error within a tolerance interval. Theorem \ref{theorem:2} further provides theoretical analysis for the superiority of our PPAD over conventional direct denoising strategies. For brevity, detailed derivations are deferred to the Appendix \ref{Theoretical Proof}.

\begin{theorem}[See the proof in Appendix~\ref{proof:error_control}]\label{theorem:1}
Consider the reverse denoising process of DDIM under the following conditions. 1) The noise prediction model $\mathcal{M}_\theta^{\mathrm{DM}}(\mathbf{x}_t,t,\mathbf{c})$ exhibits bounded prediction error: $\forall t$, $\exists\delta>0$ such that $\|\mathcal{M}_\theta^{\mathrm{DM}}(\mathbf{x}_t,t,\mathbf{c}) - \mathcal{M}_{\mathrm{GT}}^{\mathrm{DM}}(\mathbf{x}^*_t,t,\mathbf{c})\| \leq \delta$ where $\mathcal{M}_{\mathrm{GT}}^{\mathrm{DM}}(\mathbf{x}^*_t,t,\mathbf{c})$ denotes the ground truth noise.
2) There exists a threshold $\mathrm{SNR_{min}} > 0$ satisfying $\mathrm{SNR}(t)= \frac{\bar{\alpha}_t}{1-\bar{\alpha}_t} \geq \mathrm{SNR_{min}}$, $ \forall t$. Then, the cumulative error $E_{T}$ in the constrained reverse process with $\mathrm{SNR}(t)\geq\mathrm{SNR}_{\min}$ is bounded by
\begin{equation}
E_T \leq C\cdot\delta\cdot\sum_{t=1}^T\gamma_t,\quad \text{where } \gamma_t = \sqrt{\frac{1-\bar{\alpha}_t}{\bar{\alpha}_t}} \leq \sqrt{\frac{1}{\mathrm{SNR_{min}}}}
\label{eq:Effect_end}
\end{equation}
and $C$ is a constant.

\end{theorem}

\begin{theorem}[See the proof in Appendix~\ref{proof:denoising}]\label{theorem:2}
Assume the enhanced prompt $\widetilde{\mathbf{c}}_t$ contains more accurate information than prompt $\mathbf{c}$, i.e., $ \left\|\mathcal{M}_\theta^{\mathrm{DM}}(\mathbf{x}_t,t,\tilde{\mathbf{c}})-\epsilon\right\|\ll\left\|\mathcal{M}_\theta^{\mathrm{DM}}(\mathbf{x}_t,t,\mathbf{c})-\epsilon\right\| $. The PPAD denoising formulation can be written as 
\begin{equation}
\mathbf{x}_{t-2}=\underbrace{\eta_1\mathbf{x}_{t-1}}_{\text{\rm{Input}}}+\underbrace{\eta_2\mathcal{M}_\theta^{\mathrm{DM}}(\tilde{\mathbf{x}}_{t-1},t-1,\mathbf{c})}_{\text{\rm{Error Correction}}}+\underbrace{\eta_3\mathcal{M}_\theta^{\mathrm{DM}}(\tilde{\mathbf{x}}_t,t,\tilde{\mathbf{c}}_t)}_{\text{\rm{Semantic Enhancement}}}+\underbrace{\eta_4\epsilon_t}_{\text{\rm{Controllable Noise}}}
\label{eq:them2}
\end{equation}

where $\eta_1$, $\eta_2$, $\eta_3$ and $\eta_4$ are constants. 
When compared with the direct denoising baseline $\mathbf{x}_{t-2}=\eta_1\mathbf{x}_{t-1}+\eta_2\mathcal{M}_\theta^{\mathrm{DM}}(\mathbf{x}_{t-1},t-1,\mathbf{c})$, we can prove that the additional introduction of error correction term and semantic enhancement term can help the denoising error $\delta_{\mathrm{PPAD}}$ of PPAD to be lower than that of the direct denoising method $\delta_{\mathrm{direct}}$.

\end{theorem}
\section{Experiments}
\label{sec:experiments}

We conducted experiments to evaluate the effectiveness and generalizability of \method{}. More details and results are provided in the appendix. 

\subsection{Experimental Setup}

\noindent \textbf{Benchmark Datasets.}
We evaluate our proposed \method{} on \texttt{\textit{Drawbench}}, \texttt{\textit{Pick-a-Pic}}, as well as multiple subcategories within the benchmarks.

\noindent \textbf{Comparison Baselines.}
To verify the applicability, we do experiments based on some widely used baselines including (a) diffusion base model \texttt{\textit{Hunyuan-DiT}}~\cite{li2024hunyuan}, \texttt{\textit{PixArt-Sigma}}~\cite{chen2024pixart}, \texttt{\textit{SDXL}}~\cite{podell2023sdxl}, \texttt{\textit{SD1.5}}~\cite{rombach2021highresolution}. 
(b) diffusion correction framework \textit{LGD}~\cite{black2023diffusiondpo}, \textit{ZZD}~\cite{bai2024zigzag}.

\noindent \textbf{Evaluation Metrics.}
We use the widely adopted \texttt{\textit{CLIP Score}} (CLIP), \texttt{\textit{Pick Score}} (Pick), \texttt{\textit{Image Reward}} (IR), \texttt{\textit{HPSv2 Score}} (HPSv2), and \texttt{\textit{AES Score}} (AES) as the metrics. 
They respectively reflect several key aspects commonly used in text-to-image generation: semantic alignment (CLIP, Pick), subjective preference modeling (Pick, IR), and visual structure quality (HPSv2 and AES).

\noindent \textbf{Implementation Details.}
We conduct all experiments on RTX 3090 GPUs. The sampling steps $T$ are set to 50. $t_s$ is set to 0.2$T$, $t_e$ is set to 0.8$T$, and $\Delta = 5$. We adopt Qwen2.5-VL-7B~\cite{bai2025qwen25vl} as the MLLM responsible for correcting semantic information during the diffusion process.

\subsection{Experimental Results}
Except for Table~\ref{tab:main_table_lmm}, all results of PPAD are evaluated under the inference-only setting.

\subsubsection{Main Results}
Table~\ref{tab:main_result} presents the performance comparison based on pre-trained diffusion models under a pure \emph{inference-only} setting, without any additional training. All results—including VD, ZZD, and PPAD—are evaluated solely during inference. \method{} demonstrates superior performance compared to existing methods under this setting. Table~\ref{tab:main_table_lmm} shows the comparison after incorporating post-training, where both LGD and our method are applied on top of pre-trained diffusion models.

We summarize the key findings as follows:
(i) Under the pure inference setting, ZZD significantly improves performance over VD, yet it remains inferior to our method. This highlights that having a well-defined correction direction and interpretable semantic knowledge can substantially enhance text-to-image generation quality.
(ii) In the post-training setting, LGD generally serves as the strongest baseline. Nevertheless, our method outperforms even the best baseline in most cases, demonstrating the critical role of progressive correction in improving text-to-image generation performance.

\begin{table*}[ht]
\centering
\vspace{-0.2cm}
\caption{The inference-only performance of \method{} on DrawBench and Pick-a-Pic.}
\label{tab:main_result}
\vspace{-0.25cm}
\setlength{\tabcolsep}{3.5pt}
\renewcommand{\arraystretch}{0.86}
\resizebox{\textwidth}{!}{
\begin{tabular}{ll|ccccc|ccccc}
\toprule[2pt]
\multirow{2}{*}{\textbf{Model}} & \multirow{2}{*}{\textbf{Method}} 
& \multicolumn{5}{c|}{\textbf{DrawBench}} 
& \multicolumn{5}{c}{\textbf{Pick-a-Pic}} \\
\cmidrule(lr){3-7} \cmidrule(lr){8-12}
& & \textbf{CLIP} $\uparrow$ & \textbf{Pick} $\uparrow$ & \textbf{IR} $\uparrow$ & \textbf{HPSv2} $\uparrow$ & \textbf{AES} $\uparrow$ 
  & \textbf{CLIP} $\uparrow$ & \textbf{Pick} $\uparrow$ & \textbf{IR} $\uparrow$ & \textbf{HPSv2} $\uparrow$ & \textbf{AES} $\uparrow$ \\
\midrule
\midrule

\multirow{3}{*}{\textbf{Hunyuan-DiT}} 
& VD         & 28.9649 & 22.3054 & 0.8620 & 0.2874 & 5.8216 & 28.7058 & 22.0795 & 0.9941 & 0.2961 & 6.2254 \\
& ZZD           & 29.3538 & 22.3194 & 0.9390 & 0.2847 & 5.8232 & \textbf{29.0062} & 22.0446 & \textbf{0.9971} & 0.2927 & 6.2313 \\
\cmidrule{2-12}
& \cellcolor{blue!5}\method{} (Ours) & \cellcolor{blue!5}\textbf{29.4077} & \cellcolor{blue!5}\textbf{22.4001} & \cellcolor{blue!5}\textbf{0.9463} & \cellcolor{blue!5}\textbf{0.2908} & \cellcolor{blue!5}\textbf{5.8357} & \cellcolor{blue!5}28.7970 & \cellcolor{blue!5}\textbf{22.1038} & \cellcolor{blue!5}0.9969 & \cellcolor{blue!5}\textbf{0.2964} & \cellcolor{blue!5}\textbf{6.2314} \\
\midrule

\multirow{3}{*}{\textbf{SD1.5}} 
& VD         & 28.1152 & 21.0822 & -0.0208 & 0.2432 & 5.2010 & 28.5090 & 20.4871 & 0.0290 & 0.2467 & 5.4103 \\
& ZZD           & 28.5156 & 21.1293 & 0.0996  & 0.2489 & 5.2176 & 28.6023 & 20.5315 & 0.0907 & \textbf{0.2507} & 5.4213 \\
\cmidrule{2-12}
& \cellcolor{blue!5}\method{} (Ours) & \cellcolor{blue!5}\textbf{28.7690} & \cellcolor{blue!5}\textbf{21.1830} & \cellcolor{blue!5}\textbf{0.1097} & \cellcolor{blue!5}\textbf{0.2490} & \cellcolor{blue!5}\textbf{5.2844} & \cellcolor{blue!5}\textbf{28.6613} & \cellcolor{blue!5}\textbf{20.5602} & \cellcolor{blue!5}\textbf{0.0920} & \cellcolor{blue!5}0.2500 & \cellcolor{blue!5}\textbf{5.4464} \\
\midrule

\multirow{3}{*}{\textbf{PixArt-Sigma}} 
& VD   & 28.4124 & 22.1120 & 0.5060 & 0.2527 & 5.9087 & 28.6577 & 21.8092 & 0.6948 & 0.2676 & \textbf{6.1713} \\
& ZZD     & \textbf{28.9967} & 22.0180 & 0.6131 & 0.2595 & 5.7889 & \textbf{29.3186} & 21.7003 & 0.7945 & 0.2720 & 6.0708 \\
\cmidrule{2-12}
& \cellcolor{blue!5}\method{} (Ours) 
& \cellcolor{blue!5}28.8467 & \cellcolor{blue!5}\textbf{22.3278} & \cellcolor{blue!5}\textbf{0.7525} & \cellcolor{blue!5}\textbf{0.2793} & \cellcolor{blue!5}\textbf{5.9433} 
& \cellcolor{blue!5}28.9682 & \cellcolor{blue!5}\textbf{22.0168} & \cellcolor{blue!5}\textbf{0.8647} & \cellcolor{blue!5}\textbf{0.2817} & \cellcolor{blue!5}6.1684 \\

\bottomrule[2pt]
\end{tabular}
}
\vspace{-0.5cm}
\end{table*}
\begin{table*}[ht]
\centering
\caption{The post-training performance of \method{} on DrawBench and Pick-a-Pic.}
\label{tab:main_table_lmm}
\vspace{-0.25cm}
\renewcommand{\arraystretch}{0.86}
\resizebox{\textwidth}{!}{
\begin{tabular}{ll|ccccc|ccccc}
\toprule[2pt]
\multirow{2}{*}{\textbf{Model}} & \multirow{2}{*}{\textbf{Method}}
& \multicolumn{5}{c|}{\textbf{DrawBench}}
& \multicolumn{5}{c}{\textbf{Pick-a-Pic}}\\
\cmidrule(lr){3-7}\cmidrule(lr){8-12}
& & \textbf{CLIP}$\uparrow$ & \textbf{Pick}$\uparrow$ & \textbf{IR}$\uparrow$ & \textbf{HPSv2}$\uparrow$ & \textbf{AES}$\uparrow$
  & \textbf{CLIP}$\uparrow$ & \textbf{Pick}$\uparrow$ & \textbf{IR}$\uparrow$ & \textbf{HPSv2}$\uparrow$ & \textbf{AES}$\uparrow$\\
\midrule\midrule
\multirow{4}{*}{\textbf{SDXL}}
& VD & 29.1770 & 22.2266 & 0.5026 & 0.2668 & 5.6187 & 29.3724 & 21.8772 & 0.6141 & 0.2798 & 6.0428 \\
& LGD & 29.7100 & 22.5270 & \textbf{0.8484} & 0.2805 & 5.6482 & \textbf{30.1847} & 22.4207 & 0.9360 & 0.2986 & 6.0358 \\
& ZZD           & 29.2946 & 22.3337 & 0.5827 & 0.2742 & 5.6217 & 29.6694 & 22.1136 & 0.9009 & 0.2976 & 6.0393 \\
\cmidrule{2-12}
& \cellcolor{blue!5}\method{} (Ours)
& \cellcolor{blue!5}\textbf{29.9322} & \cellcolor{blue!5}\textbf{22.6502} & \cellcolor{blue!5}0.8402 & \cellcolor{blue!5}\textbf{0.2856} & \cellcolor{blue!5}\textbf{5.6824}
& \cellcolor{blue!5}30.1580 & \cellcolor{blue!5}\textbf{22.4538} & \cellcolor{blue!5}\textbf{0.9730} & \cellcolor{blue!5}\textbf{0.3008} & \cellcolor{blue!5}\textbf{6.0542}\\
\bottomrule[2pt]
\end{tabular}}
\vspace{-0.3cm}
\end{table*}

\subsubsection{Category-wise Performance}
\begin{table*}[!t]
\centering
\caption{HPSv2 Score comparison across categories.}
\label{tab:main_detail}
\vspace{-0.25cm}
\setlength{\tabcolsep}{2.5pt}
\renewcommand{\arraystretch}{0.95}
\resizebox{\textwidth}{!}{
\begin{tabular}{lccccccccccc}
\toprule[2pt]
\textbf{Method} & \textbf{Colors} & \textbf{Conflicting} & \textbf{Counting} & \textbf{DALL-E} & \textbf{Descriptions} & \textbf{Marcus et al.} & \textbf{Misspellings} & \textbf{Positional} & \textbf{Rare Words} & \textbf{Reddit} & \textbf{Text} \\
\midrule
\midrule
VD & 0.3018 & 0.3210 & 0.3135 & 0.2896 & \textbf{0.2693} & \textbf{0.2976} & 0.2051 & 0.2991 & \textbf{0.2100} & 0.3088 & 0.2566 \\
Zigzag   & 0.2969 & 0.3152 & 0.3091 & 0.2896 & 0.2593 & 0.2830 & 0.2030 & 0.3081 & 0.2053 & 0.3115 & 0.2495 \\
\midrule
\rowcolor{blue!5}\method{} (Ours)    & \textbf{0.3044} & \textbf{0.3284} & \textbf{0.3186} & \textbf{0.2913} & 0.2676 & 0.2925 & \textbf{0.2078} & \textbf{0.3103} & 0.2031 & \textbf{0.3130} & \textbf{0.2625} \\
\bottomrule[2pt]
\end{tabular}
}
\vspace{-0.6cm}
\end{table*}
To better understand the effectiveness of our proposed \method{}, we conduct a fine-grained analysis of its performance across different prompt categories under the HPSv2 metric, which emphasizes perceptual quality, object fidelity, and spatial structure.
We observe that \method{} achieves notable improvements over baselines in categories such as \textit{Colors}, \textit{Conflicting}, \textit{Counting}, \textit{DALL-E}, \textit{Misspellings}, \textit{Positional}, \textit{Reddit}, and \textit{Text}. These categories typically involve explicit visual attributes (e.g., color, count, position) or compositional reasoning with moderate prompt complexity. 
The \textit{LSG} renders clearer mid-process snapshots, enabling the \textit{SCC-CPS pipeline} to effectively detect semantic mismatches and refine vague descriptions. Additionally, the \textit{PPA} mitigates error accumulation and guides the model toward semantically faithful outputs. 
In contrast, our method shows relatively lower performance in categories such as \textit{Descriptions}, \textit{Marcus et al.}, and \textit{Rare Words}. These categories pose greater challenges due to complex or ill-defined semantics. In \textit{Descriptions}, prompts are often long and detail-heavy, which increases the difficulty of generating meaningful and structurally consistent corrections via MLLM. The resulting refined prompt may inadvertently oversimplify or misrepresent the original intent. For \textit{Marcus et al.} prompts, which include abstract or paradoxical prompts, the MLLM struggles to produce actionable guidance, limiting the effectiveness of semantic correction. Similarly, in the \textit{Rare Words} category, the MLLM may lack sufficient visual grounding for obscure tokens, weakening its ability to inject meaningful semantic feedback.

\subsubsection{Visualization Results}
\label{sec:vis_case}
\begin{figure}[!ht]
    \centering
    \vspace{-0.35cm}
    \includegraphics[width=0.96\textwidth]{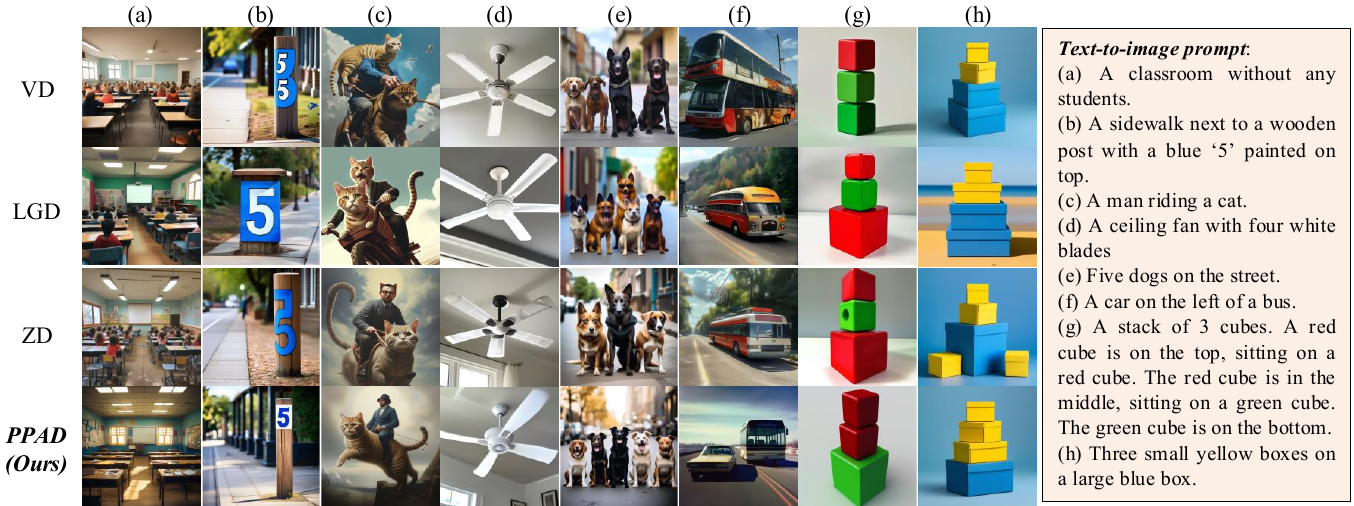}
    \vspace{-0.1cm}
    \caption{Comparison of generated images.}
    \label{fig:generation_case1}
    \vspace{-0.5cm}
\end{figure}

\noindent \textbf{Generation Comparisons.}
Figure~\ref{fig:generation_case1} presents a qualitative comparison between our proposed PPAD and the baselines on several representative prompts. These prompts span categories such as \textit{Negation}, \textit{Text}, \textit{Conflicting}, \textit{Counting}, \textit{Positional}, and multi-category composite prompts.
\uline{(a) Negation.} It requires the model to understand negation. All baselines fail by generating students in the scene. In contrast, PPAD successfully renders an empty classroom.
\uline{(b) Text.} Only PPAD successfully produces a clean, readable digit in the correct location. None of the baselines correctly understood this text, which includes positional relationships and numerical information and so on.
\uline{(c) Conflicting Semantics.} Most baselines reverse the subject–object roles. PPAD uniquely renders the correct configuration.
\uline{(d, e) Counting.} These examples are challenging for generative models. Existing baselines often produce inaccurate object counts or incorrect backgrounds, while PPAD consistently generates the correct number of objects and ensures accurate background details.
\uline{(f) Positional Relation.} It tests spatial layout understanding. While other methods often swap or overlap entities, PPAD correctly captures the left–right relationship.
\uline{(g, h) Spatial Composition and Color.} Only PPAD produces the correct order and color while maintaining object coherence. Other methods fail to align both color and spatial arrangement.
In summary, PPAD effectively enables real-time semantic correction during inference, achieving a higher degree of semantic alignment.

\begin{figure}[t]
    \centering
    \includegraphics[width=0.96\textwidth]{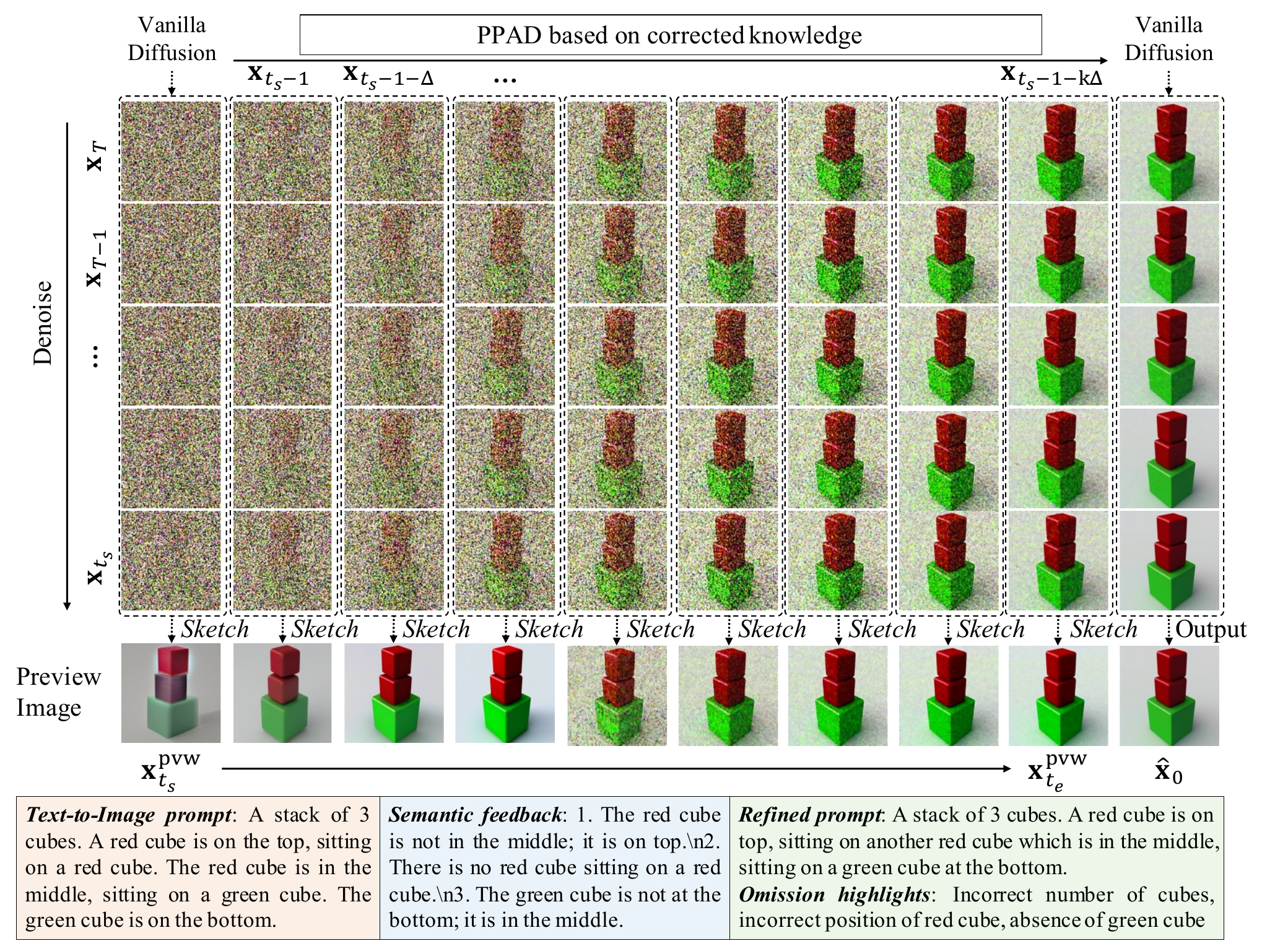}
    \vspace{-0.2cm}
    \caption{Case of denoising path.}
    \label{fig:path_case_1}
    \vspace{-0.6cm}
\end{figure}

\noindent \textbf{Denoising Path.} Figure~\ref{fig:path_case_1} visualizes a case of structured object hallucination and how PPAD corrects the denoising trajectory. The text-to-image prompt describes a stack of three cubes in a specific top–middle–bottom order. However, the initial sketch $\mathbf{x}^{\text{pvw}}_{t_s}$ shows a red cube directly stacked on a green one, missing the middle cube entirely. The MLLM evaluates this preview and provides fine-grained semantic feedback, revealing incorrect object count and position.
Based on this semantic feedback, a refined prompt and a set of omission highlights are generated. These corrections guide the subsequent denoising stages from $\mathbf{x}_{t_s}$ to $\mathbf{x}_{t_e}$, effectively addressing hallucination errors such as missing middle cubes and misplaced color stacking. After $\mathbf{x}_{t_e}$, the model reverts to vanilla diffusion for final synthesis. Compared to the initial path, the corrected trajectory restores the intended spatial and structural configuration of the cubes.

\subsubsection{Ablation Study}
\label{sec:ablation_study}
Table~\ref{tab:ablation_module} presents the performance of our method with different module combinations based on Hunyuan-DiT on DrawBench. The first row corresponds to Vanilla Diffusion (VD). Interestingly, enabling only SCK (second row) leads to performance drops on most metrics. This is because intermediate latent states are typically noisy and ambiguous, and relying on MLLM to produce SCK at such stages can introduce inaccurate or even misleading semantic feedback. However, when SCK is combined with LKG (third row), overall performance improves, indicating that LKG’s adaptive previewing produces clearer intermediate images that enable more reliable semantic evaluation by SCK.
\begin{wraptable}{r}{0.5\textwidth}
\vspace{-0.3cm}
\caption{Module ablation study.}
\label{tab:ablation_module}
\vspace{-0.25cm}
\centering
\setlength{\tabcolsep}{3pt}
\renewcommand{\arraystretch}{1}
\resizebox{0.5\textwidth}{!}{
\begin{tabular}{ccc|ccccc}
\toprule[2pt]
SCK & LKG & PPA & CLIP~($\uparrow$) & Pick~($\uparrow$) & IR~($\uparrow$) & HPSv2~($\uparrow$) & AES~($\uparrow$) \\
\midrule\midrule
\XSolidBrush & \XSolidBrush & \XSolidBrush & \uline{28.9649} & \uline{22.3054} & 0.8620 & \uline{0.2874} & 5.8216 \\
\Checkmark & \XSolidBrush & \XSolidBrush & 28.1228 & 22.2304 & 0.7653 & 0.2861 & \textbf{5.8974} \\
\Checkmark & \Checkmark & \XSolidBrush & 28.9386 & 22.3037 & \uline{0.9167} & 0.2859 & 5.8031 \\
\rowcolor{blue!5}\Checkmark & \Checkmark & \Checkmark & \textbf{29.4077} & \textbf{22.4001} & \textbf{0.9463} & \textbf{0.2908} & \uline{5.8357} \\
\bottomrule[2pt]
\end{tabular}
}
\vspace{-0.3cm}
\end{wraptable}
When all three modules are enabled (last row), our method achieves the best performance across all metrics. This demonstrates that PPA is beneficial for injecting and applying SCK to refine the diffusion process. In this configuration, SCK provides more effective semantic correction, LKG ensures high-quality intermediate inputs, and PPA enables SCK to better guide the denoising trajectory.

\section{Conclusion}
\label{sec:conclusion}
We propose PPAD, a novel framework that for the first time enables semantic feedback during the denoising process by leveraging MLLMs. Unlike prior methods that rely on static prompts or post-hoc scoring, PPAD actively inspects and corrects semantic deviations throughout generation, significantly improving alignment and compositional quality.
Beyond performance gains, PPAD establishes a bridge between MLLMs and diffusion models, demonstrating the feasibility of collaborative reasoning and generation. We believe this work opens new directions for integrating language understanding into controllable and interpretable visual synthesis.
\clearpage
\section*{Limitation}
\label{sec:limitation}
While PPAD introduces semantic inspection and correction into the denoising process and demonstrates strong effectiveness, it still has limitations. In cases where the MLLM underperforms—such as with rare or underrepresented nouns and concepts (e.g., obscure object names or unconventional compositional instructions)—the improvements become less significant and may even lead to performance degradation. This is mainly due to the limited ability of current MLLMs to understand and ground unfamiliar semantics.
\section*{Broader Impact}
\label{sec:broader_impact}
PPAD marks a step toward more semantically grounded and interpretable generative models by tightly integrating multimodal reasoning into the generation process. By embedding MLLM-based feedback loops into diffusion inference, PPAD enables fine-grained semantic alignment and progressive correction—capabilities that are critical in high-stakes applications such as medical imaging, scientific visualization, and education. In addition, its modular design offers a general-purpose interface for injecting structured linguistic supervision into generative systems. We believe this direction opens up new opportunities for human-in-the-loop generation, semantic safety auditing, and adaptive control of multimodal AI systems.
\bibliographystyle{unsrt}
\bibliography{reference}
\clearpage
\appendix
\section{Appendix}
\label{sec:appendix}

\subsection{Pseudo Code}

Algorithm~\ref{alg:ppad} shows the inference process of our proposed PPAD, which integrates semantic feedback into the denoising steps. 
Specifically, at certain timesteps within a predefined correction interval, the model conditionally triggers a semantic correction process if the signal-to-noise ratio exceeds a given threshold. 
The semantic discrepancy between the generated image and the prompt is evaluated by a multimodal large language model (MLLM). 
If the discrepancy remains above a stopping threshold, the MLLM generates a rewritten prompt, which is encoded and used to refine the denoising in subsequent steps.

\begin{algorithm}[!ht]
\caption{PPAD: MLLM Semantic-Corrected Ping–Pong–Ahead Diffusion}
\label{alg:ppad}
\DontPrintSemicolon
\KwIn{Prompt $\mathbf{p}$; text encoder $\mathtt{E}$; diffusion model $\mathcal{M}^{\textsc{dm}}_{\theta}$;\\
\hspace{1.95em} total steps $T$; correction interval $[t_e,t_s]$; stride $\Delta$;\\
\hspace{1.95em} SNR threshold $\gamma$; early-stop threshold $\tau_{\text{stop}}$.}
\KwOut{Final image $\hat{\mathbf{x}}_{0}$.}

$\mathbf{c} \leftarrow \mathtt{E}(\mathbf{p})$ \tcp*[r]{text condition}
Sample $\mathbf{x}_{T} \sim \mathcal{N}(\mathbf{0},\mathbf{I})$\;
\For{$t = T,\,T-1,\,\dots,\,1$}{
    $k \leftarrow \mathbbm{1}[\mathrm{SNR}(t)>\gamma] + 1$\;
    $\mathbf{x}^{\text{pvw}}_{t} \leftarrow 
        \mathcal{M}^{\textsc{dm}}_{\theta}(\mathbf{x}_{t-k},\,t-k,\,\mathbf{c})$\;
    \If{$t \in \{t_s-k\Delta,\,t_s-(k+1)\Delta,\dots,t_e\}$}{
        $(s_{t},S_{t}) \leftarrow
            \mathcal{M}^{\textsc{mllm}}(\mathbf{x}^{\text{pvw}}_{t},\mathbf{p})$\;
        \lIf{$s_{t} \ge \tau_{\text{stop}}$}{\textbf{continue}}
        $(s^{+}_{t},s^{-}_{t}) \leftarrow 
            \mathcal{M}^{\textsc{mllm}}\texttt{.rewrite}(S_{t})$\;
        $\mathbf{c}^{+}_{t} \leftarrow \mathtt{E}(s^{+}_{t}),\;
         \mathbf{c}^{-}_{t} \leftarrow \mathtt{E}(s^{-}_{t})$\;
        $\tilde{\mathbf{c}}_{t} \leftarrow \mathbf{c}^{+}_{t}\,\Vert\,\mathbf{c}^{-}_{t}$\;
        $\tilde{\mathbf{x}}_{t} \leftarrow 
            \widetilde{\mathcal{F}}_{t-1\to t}(\mathbf{x}_{t-1})$ \tcp*[r]{Ping}
        $\tilde{\mathbf{x}}_{t-1} \leftarrow 
            \mathcal{F}_{t\to t-1}(\tilde{\mathbf{x}}_{t},\tilde{\mathbf{c}}_{t})$ \tcp*[r]{Pong}
        $\mathbf{x}_{t-2} \leftarrow 
            \mathcal{F}_{t-1\to t-2}(\tilde{\mathbf{x}}_{t-1},\mathbf{c})$ \tcp*[r]{Ahead}
        $t \leftarrow t-1$ \tcp*[r]{extra step consumed}
    }
    \Else{
        $\mathbf{x}_{t-1} \leftarrow 
            \mathcal{F}_{t\to t-1}(\mathbf{x}_{t},\mathbf{c})$\;
    }
}
\Return $\hat{\mathbf{x}}_{0} = \mathbf{x}_{0}$\;
\end{algorithm}

Listing~\ref{code:ppad} provides the Python implementation of auxiliary modules used in this process. 
The \texttt{\small analyze\_mismatches} function provides semantic feedback according to the generated image and the original prompt, serving as the basis for subsequent refinement.
The \texttt{\small get\_refined\_prompt} function synthesizes an enhanced version of the original prompt to better guide image generation, based on semantic mismatches detected in the current image. 
The \texttt{\small get\_omission\_highlight} function identifies not only undesirable visual features but also semantically misaligned elements in the current image, helping the diffusion model avoid repeating these issues in subsequent generations.

\begin{lstlisting}[
style=pythonstyle, 
caption={Prompt diagnosis and enhancement functions.}, 
label={code:ppad},
belowcaptionskip=-0.2cm
]
def analyze_mismatches(self, image_path, original_prompt, return_ids=False):
    """Analyze mismatches between the image and the original prompt"""
    question_text = f"""
    Analyze the image and identify all mismatches with the original prompt.

    Original prompt: "{original_prompt}"

    Instructions:
    1. List ALL elements from the prompt that are missing in the image.
    2. List ALL elements from the prompt that appear incorrectly (wrong quantity, appearance, position, etc.).
    3. Be precise and specific in your analysis.

    Format your response as a numbered list of issues ONLY.
    """
    return self.ask_vlm(image_path, question_text, return_ids)

def get_refined_prompt(self, image_path, original_prompt, diagnosis, return_ids=False):
    """Generate an refined prompt to better match the original intent"""
    question_text = f"""
    You are an expert prompt engineer for image generation models.

    Original prompt: "{original_prompt}"

    Issues with the current image:
    {diagnosis}

    Instructions:
    1. Create an improved prompt that will help the image generation model better match the original intention.
    2. Add specific details, emphasis, or clarifications to address the identified issues.
    3. Maintain the core idea and style of the original prompt - do not add unrelated concepts.
    4. The goal is to get an image closer to what was originally intended.
    5. Use techniques like emphasis words, specific quantities, spatial relationships, or other details as needed.

    Return only one well-structured, fluent sentence without any explanations.
    """
    return self.ask_vlm(image_path, question_text, return_ids)

def get_omission_highlight(self, image_path, original_prompt, diagnosis, return_ids=False):
    """Generate a omission highlight to eliminate unwanted elements"""
    question_text = f"""
    Based on the original prompt and analysis of the current image, list elements that should be avoided.

    Original prompt: "{original_prompt}"

    Current image issues:
    {diagnosis}

    Instructions:
    1. List quality issues to avoid
    2. DO NOT include any objects from the prompt.

    Return only comma-separated quality terms.
    """
    return self.ask_vlm(image_path, question_text, return_ids)

\end{lstlisting}

\subsection{Theoretical Proof}\label{Theoretical Proof}
\subsubsection{Proof of Theorem~\ref{theorem:1}}
\label{proof:error_control}

We demonstrate that enforcing the SNR threshold $\mathrm{SNR}(t)\geq\mathrm{SNR}_{\min}$ ensures controlled error propagation through following key steps.

Let $\mathbf{x}^*_t$ denote the ideal state under perfect noise prediction and $e_t=\|\mathbf{x}_t-\mathbf{x}^*_t\|_2$ represent the error per step. From the DDIM update rule
\begin{equation}
\mathbf{x}_{t-1} = \sqrt{\frac{\bar{\alpha}_{t-1}}{\bar{\alpha}_t}}\mathbf{x}_t + \left( \sqrt{1-\bar{\alpha}_{t-1}} - \sqrt{\frac{\bar{\alpha}_{t-1}}{\bar{\alpha}_t}(1-\bar{\alpha}_t)} \right)\mathcal{M}_\theta^{\mathrm{DM}}(\mathbf{x}_t,t,\mathbf{c}).
\label{eq:ddim_update}
\end{equation}
The corresponding ideal state update is (assuming the model predicts no error)
\begin{equation}
\mathbf{x}_{t-1}^*=\sqrt{\frac{\bar{\alpha}_{t-1}}{\bar{\alpha}_t}}\mathbf{x}_t^*+\left(\sqrt{1-\bar{\alpha}_{t-1}}-\sqrt{\frac{\bar{\alpha}_{t-1}}{\bar{\alpha}_t}(1-\bar{\alpha}_t)}\right)\mathcal{M}_{\mathrm{GT}}^{\mathrm{DM}}(\mathbf{x}^*_t,t,\mathbf{c}).
\label{eq:idea_update}
\end{equation}
The error recursion relationship between the actual and ideal states is:
\begin{align}
\mathbf{e}_{t-1}&=\|\mathbf{x}_{t-1}-\mathbf{x}_{t-1}^*\| \\
  &\leq\sqrt{\frac{\bar{\alpha}_{t-1}}{\bar{\alpha}_t}}\cdot\|\mathbf{x}_t-\mathbf{x}_t^*\|+\left\|\left(\sqrt{1-\bar{\alpha}_{t-1}}-\sqrt{\frac{\bar{\alpha}_{t-1}}{\bar{\alpha}_t}(1-\bar{\alpha}_t)}\right)(\mathcal{M}_\theta^{\mathrm{DM}}-\mathcal{M}_{\mathrm{GT}}^{\mathrm{DM}})\right\|.
  \label{eq:idea_act}
\end{align}

Let the model prediction error be $\|\mathcal{M}_\theta^{\mathrm{DM}}(\mathbf{x}_t,t,\mathbf{c}) - \mathcal{M}_{\mathrm{GT}}^{\mathrm{DM}}(\mathbf{x}^*_t,t,\mathbf{c})\| \leq \delta$, and define the coefficients
\begin{equation}
\gamma_t=\left|\sqrt{1-\bar{\alpha}_{t-1}}-\sqrt{\frac{\bar{\alpha}_{t-1}}{\bar{\alpha}_t}(1-\bar{\alpha}_t)}\right|.
\label{eq:coefficients}
\end{equation}
Then the error recursion inequality is simplified to
\begin{equation}
e_{t-1}\leq\sqrt{\frac{\bar{\alpha}_{t-1}}{\bar{\alpha}_t}}\cdot e_t+\gamma_t\delta.
\label{eq:simplified}
\end{equation}

Through algebraic manipulation, we derive
\begin{equation}
\gamma_t = \sqrt{\frac{1-\bar{\alpha}_t}{\bar{\alpha}_t}} \left( 1 - \sqrt{ \frac{\bar{\alpha}_{t-1}(1-\bar{\alpha}_t)}{\bar{\alpha}_t(1-\bar{\alpha}_{t-1})} } \right).
\label{eq:gamma_transform}
\end{equation}
Under the $\bar{\alpha}_{t}\geq\frac{\mathrm{SNR}_{\min}}{1+\mathrm{SNR}_{\min}}$, this simplifies to
\begin{equation}
\gamma_t \leq \sqrt{\frac{1-\bar{\alpha}_t}{\bar{\alpha}_t}} \leq \sqrt{\frac{1}{\mathrm{SNR_{min}}}}.
\label{eq:gamma_bound}
\end{equation}
This means that the coefficient $\gamma_{t}$ of the error at each step is strictly limited.
With initial condition $e_{T}=0$, backward recursion yields
\begin{equation}
e_0 \leq \delta\sum_{t=1}^T\gamma_t\prod_{s=1}^{t-1}\sqrt{\frac{\bar{\alpha}_s}{\bar{\alpha}_{s+1}}}.
\label{eq:error_expansion}
\end{equation}
Due to $\sqrt{\bar{\alpha}_s/\bar{\alpha}_{s+1}}\leq1$ ($\bar{\alpha}_s\leq\bar{\alpha}_{s+1}$), the product term is controlled by exponential decay, and the final cumulative error is bounded by
\begin{equation}
E_T = e_0 \leq C\delta\sum_{t=1}^T\gamma_t \leq C\delta T\sqrt{\frac{1}{\mathrm{SNR_{min}}}},
\label{eq:final_bound}
\end{equation}
where $C$ absorbs constants from the recursive bounds.

When $\exists t$ with $\bar{\alpha}_t\to0$, the growth factor $\gamma_t\approx\sqrt{1/\bar{\alpha}_t}\to\infty$ causes divergence in \eqref{eq:final_bound}. The SNR constraint prevents this pathological case by maintaining uniformly across timesteps.

\subsubsection{Proof of Theorem~\ref{theorem:2}}
\label{proof:denoising}
Direct (vanilla) denoising method
\begin{equation}
\mathbf{x}_{t-2}=\sqrt{\frac{\bar{\alpha}_{t-2}}{\bar{\alpha}_{t-1}}}\mathbf{x}_{t-1}+\left(\sqrt{1-\bar{\alpha}_{t-2}}-\sqrt{\frac{\bar{\alpha}_{t-2}(1-\bar{\alpha}_{t-1})}{\bar{\alpha}_{t-1}}}\right)\mathcal{M}_\theta^{\mathrm{DM}}(\mathbf{x}_{t-1},t-1,\mathbf{c}).
\label{eq:Direct denoising method}
\end{equation}

Ping-Pong-Ahead Diffusion (PPAD)

\begin{align}
\textbf{Ping} :\;&
\tilde{\mathbf{x}}_t=\sqrt{\frac{\bar{\alpha}_t}{\bar{\alpha}_{t-1}}}\mathbf{x}_{t-1}+\sqrt{1-\bar{\alpha}_t}\epsilon_t,
       \label{eq:ping_ddim_proof}\\
\textbf{Pong} :\;&
\tilde{\mathbf{x}}_{t-1}=\sqrt{\frac{\bar{\alpha}_{t-1}}{\bar{\alpha}_t}}\tilde{\mathbf{x}}_t+\left(\sqrt{1-\bar{\alpha}_{t-1}}-\sqrt{\frac{\bar{\alpha}_{t-1}(1-\bar{\alpha}_t)}{\bar{\alpha}_t}}\right)\mathcal{M}_\theta^{\mathrm{DM}}(\tilde{\mathbf{x}}_t,t,\tilde{\mathbf{c}}_t)
\label{eq:pong_ddpm_proof}\\
\textbf{Ahead} :\;&
\mathbf{x}_{t-2}=\sqrt{\frac{\bar{\alpha}_{t-2}}{\bar{\alpha}_{t-1}}}\tilde{\mathbf{x}}_{t-1}+\left(\sqrt{1-\bar{\alpha}_{t-2}}-\sqrt{\frac{\bar{\alpha}_{t-2}(1-\bar{\alpha}_{t-1})}{\bar{\alpha}_{t-1}}}\right)\mathcal{M}_\theta^{\mathrm{DM}}(\tilde{\mathbf{x}}_{t-1},t-1,\mathbf{c}).
\label{eq:ahead_proof}
\end{align}
Substitute Eq. (\ref{eq:ping_ddim_proof}) and Eq. (\ref{eq:pong_ddpm_proof}) into Eq. (\ref{eq:ahead_proof}) to get PPAD denoising process
\begin{align}
\mathbf{x}_{t-2}&=\sqrt{\frac{\bar{\alpha}_{t-2}}{\bar{\alpha}_{t-1}}}\left(\sqrt{\frac{\bar{\alpha}_{t-1}}{\bar{\alpha}_{t}}}\tilde{\mathbf{x}}_{t}+\left(\sqrt{1-\bar{\alpha}_{t-1}}-\sqrt{\frac{\bar{\alpha}_{t-1}(1-\bar{\alpha}_{t})}{\bar{\alpha}_{t}}}\right)\mathcal{M}_{\theta}^{\mathrm{DM}}(\tilde{\mathbf{x}}_{t},t,\tilde{\mathbf{c}}_{t})\right)\\
&+\left(\sqrt{1-\bar{\alpha}_{t-2}}-\sqrt{\frac{\bar{\alpha}_{t-2}(1-\bar{\alpha}_{t-1})}{\bar{\alpha}_{t-1}}}\right)\mathcal{M}_{\theta}^{\mathrm{DM}}(\tilde{\mathbf{x}}_{t-1},t-1,\mathbf{c}) \\
\mathbf{x}_{t-2}&=\sqrt{\bar{\alpha}_{t-1}}\Bigg(\sqrt{\bar{\alpha}_{t-1}}\Bigg(\sqrt{\bar{\alpha}_{t-1}}\mathbf{x}_{t-1}\sqrt{1-\bar{\alpha}_{t}}\mathbf{\epsilon}_{t}\Bigg)\\
&+\Bigg(\sqrt{1-\bar{\alpha}_{t-1}}-\sqrt{\frac{\bar{\alpha}_{t-1}(1-\bar{\alpha}_{t})}{\bar{\alpha}_{t}}\Bigg)}\mathcal{M}_{\theta}^{\mathrm{DM}}(\bar{\mathbf{x}}_{t},t,\tilde{\mathbf{c}}_{t})\Bigg)\\
&+\Bigg(\sqrt{1-\bar{\alpha}_{t-2}}-\sqrt{\frac{\bar{\alpha}_{t-2}(1-\bar{\alpha}_{t-1})}{\bar{\alpha}_{t-1}}}\Bigg)\mathcal{M}_{\theta}^{\mathrm{DM}}(\tilde{\mathbf{x}}_{t-1},t-1,\mathbf{c}) \\
\mathbf{x}_{t-2}&=\sqrt{\bar{\alpha}_{t-2}}\left(\mathbf{x}_{t-1}+\sqrt{\frac{\bar{\alpha}_{t-1}(1-\bar{\alpha}_t)}{\bar{\alpha}_t}}\epsilon_t +\left(\sqrt{1-\bar{\alpha}_{t-1}}-\sqrt{\frac{\bar{\alpha}_{t-1}(1-\bar{\alpha}_t)}{\bar{\alpha}_t}}\right)\mathcal{M}_\theta^{\mathrm{DM}}(\tilde{\mathbf{x}}_t,t,\tilde{\mathbf{c}}_t)\right)\\
&+\left(\sqrt{1-\bar{\alpha}_{t-2}}-\sqrt{\frac{\bar{\alpha}_{t-2}(1-\bar{\alpha}_{t-1})}{\bar{\alpha}_{t-1}}}\right)\mathcal{M}_\theta^{\mathrm{DM}}(\tilde{\mathbf{x}}_{t-1},t-1,\mathbf{c}) \\
\mathbf{x}_{t-2}&=\sqrt{\frac{\bar{\alpha}_{t-2}}{\bar{\alpha}_{t-1}}}\mathbf{x}_{t-1}+\sqrt{\frac{\bar{\alpha}_{t-2}(1-\bar{\alpha}_{t})}{\bar{\alpha}_{t}}}\epsilon_{t}\\
&+\sqrt{\frac{\bar{\alpha}_{t-2}}{\bar{\alpha}_{t-1}}}{\left(\sqrt{1-\bar{\alpha}_{t-1}}-\sqrt{\frac{\bar{\alpha}_{t-1}(1-\bar{\alpha}_{t})}{\bar{\alpha}_{t}}}\right)}\mathcal{M}_{\theta}^{\mathrm{DM}}(\tilde{\mathbf{x}}_{t},t,\tilde{\mathbf{c}}_{t})\\
&+\left(\sqrt{1-\bar{\alpha}_{t-2}}-\sqrt{\frac{\bar{\alpha}_{t-2}(1-\bar{\alpha}_{t-1})}{\bar{\alpha}_{t-1}}}\right)\mathcal{M}_{\theta}^{\mathrm{DM}}(\tilde{\mathbf{x}}_{t-1},t-1,\mathbf{c}) \\
\mathbf{x}_{t-2}&=\sqrt{\frac{\bar{\alpha}_{t-2}}{\bar{\alpha}_{t-1}}}\mathbf{x}_{t-1}+\sqrt{\frac{\bar{\alpha}_{t-2}(1-\bar{\alpha}_{t})}{\bar{\alpha}_{t}}}\epsilon_{t}\\
&+\left(\sqrt{\frac{\bar{\alpha}_{t-2}(1-\bar{\alpha}_{t-1})}{\bar{\alpha}_{t-1}}}-\sqrt{\frac{\bar{\alpha}_{t-2}(1-\bar{\alpha}_{t})}{\bar{\alpha}_{t}}}\right)\mathcal{M}_{\theta}^{\mathrm{DM}}(\tilde{\mathbf{x}}_{t},t,\tilde{\mathbf{c}}_{t})\\
&+\left(\sqrt{1-\bar{\alpha}_{t-2}}-\sqrt{\frac{\bar{\alpha}_{t-2}(1-\bar{\alpha}_{t-1})}{\bar{\alpha}_{t-1}}}\right)\mathcal{M}_{\theta}^{\mathrm{DM}}(\tilde{\mathbf{x}}_{t-1},t-1,\mathbf{c}).
\label{eq:ppad_proof}
\end{align}
To simplify, we let
\begin{align}
\eta_1&=\sqrt{\frac{\bar{\alpha}_{t-2}}{\bar{\alpha}_{t-1}}} \\
\eta_2&=\sqrt{1-\bar{\alpha}_{t-2}}-\sqrt{\frac{\bar{\alpha}_{t-2}(1-\bar{\alpha}_{t-1})}{\bar{\alpha}_{t-1}}} \\
\eta_3&=\sqrt{\frac{\bar{\alpha}_{t-2}(1-\bar{\alpha}_{t-1})}{\bar{\alpha}_{t-1}}}-\sqrt{\frac{\bar{\alpha}_{t-2}(1-\bar{\alpha}_t)}{\bar{\alpha}_t}} \\
\eta_4&=\sqrt{\frac{\bar{\alpha}_{t-2}(1-\bar{\alpha}_t)}{\bar{\alpha}_t}}.
\label{eq:eta}
\end{align}
Then, direct denoising method (Eq. (\ref{eq:Direct denoising method})) and PPAD (Eq. (\ref{eq:ppad_proof})) can be rewritten as
\begin{align}
\textbf{Direct} :\;&
\mathbf{x}_{t-2}=\underbrace{\eta_1\mathbf{x}_{t-1}}_{\text{\rm{A. Input}}}+\underbrace{\eta_2\mathcal{M}_\theta^{\mathrm{DM}}(\mathbf{x}_{t-1},t-1,\mathbf{c})}_{\text{\rm{B. Orig Error}}},
\label{eq:final_direct_proof}\\
\textbf{PPAD} :\;&
\mathbf{x}_{t-2}=\underbrace{\eta_1\mathbf{x}_{t-1}}_{\text{\rm{C. Input}}}+\underbrace{\eta_2\mathcal{M}_\theta^{\mathrm{DM}}(\tilde{\mathbf{x}}_{t-1},t-1,\mathbf{c})}_{\text{\rm{D. Error Correction}}}+\underbrace{\eta_3\mathcal{M}_\theta^{\mathrm{DM}}(\tilde{\mathbf{x}}_t,t,\tilde{\mathbf{c}}_t)}_{\text{\rm{E. Semantic Enhancement}}}+\underbrace{\eta_4\epsilon_t}_{\text{\rm{F. Controllable Noise}}}.
\label{eq:final_PPAD_proof}
\end{align}
Given the assumption that the enhanced prompt $\widetilde{\mathbf{c}}_t$ contains more accurate information than prompt $\mathbf{c}$, i.e., $ \left\|\mathcal{M}_\theta^{\mathrm{DM}}(\mathbf{x}_t,t,\tilde{\mathbf{c}})-\epsilon\right\|\ll\left\|\mathcal{M}_\theta^{\mathrm{DM}}(\mathbf{x}_t,t,\mathbf{c})-\epsilon\right\| $. We can derive that the denoising error $\delta_{E}$ in Eq. (\ref{eq:final_PPAD_proof}) is much smaller than the error $\delta_{B}$ in Eq. (\ref{eq:final_direct_proof}). i.e., $\delta_{E}\ll\delta_{B}$. Considering that the one-step forward diffusion process induces negligible error accumulation, the value of $F$ term can be ignored. Meanwhile, since $\tilde{\mathbf{x}}_{t-1}$ is the initialization noise corrected by accurate information compared to $\mathbf{x}_{t-1}$, we have $\delta_{D}<\delta_{B}$. 
Based on the above analysis, we can conclude that the denoising error $\delta_{\mathrm{PPAD}}$ of PPAD is smaller than that of the direct denoising method $\delta_{\mathrm{direct}}$. i.e. $\delta_{\mathrm{PPAD}}=(\delta_{D}+\delta_{E}) < \delta_{\mathrm{direct}}=\delta_{B}$.

\subsection{Cost Analysis} 
\begin{figure}[!ht]
    \centering
    \includegraphics[width=\textwidth]{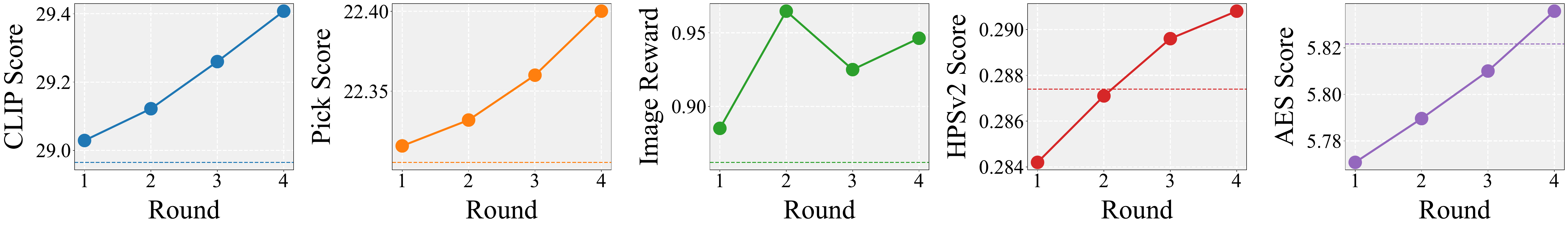}
    \caption{Performance under different rounds of MLLM invocationsr in PPAD.}
    \label{fig:cost_performance}
\end{figure}

Figure~\ref{fig:cost_performance} and Table~\ref{tab:cost_time} jointly illustrate the trade-off between performance and inference time across different rounds of semantic correction in \method{}. We compare four settings:  
\begin{itemize}[leftmargin=1.5em]
\item \textbf{PPAD-1r:} All semantic contents (\textit{consistency judgement, semantic feedback, refined prompt, omission highlights}) are generated in a single round.
\item \textbf{PPAD-2r:} Divided into two rounds: (\textit{consistency judgement + semantic feedback}), then (\textit{refined prompt, omission highlights}).
\item \textbf{PPAD-3r:} Three rounds: (\textit{consistency judgement, semantic feedback}), then (\textit{refined prompt + omission highlights}).
\item \textbf{PPAD-4r:} Each component occupies a separate round.
\end{itemize}
\begin{table}[!ht]
\centering
\setlength{\tabcolsep}{6pt}
\caption{Inference time under different rounds of MLLM invocation in PPAD.}
\label{tab:cost_time}
\resizebox{0.7\textwidth}{!}{
\begin{tabular}{lccccc}
\toprule[2pt]
\textbf{Round} & VD & PPAD-1r & PPAD-2r & PPAD-3r & PPAD-4r \\
\midrule
\textbf{Time (s)} & 122.48 & 172.53 & 189.95 & 203.18 & 210.77 \\
\bottomrule[2pt]
\end{tabular}
}
\end{table}

As shown in Table~\ref{tab:cost_time}, inference time increases gradually from 172.5s (1 round) to 210.8s (4 rounds), compared to the vanilla diffusion baseline (122.5s). However, the additional cost brings consistent and meaningful gains across all evaluation metrics (Figure~\ref{fig:cost_performance}).
These results suggest that while more rounds introduce higher cost, they also enable more refined understanding and correction of intermediate states. \textit{PPAD-2r offers a strong balance}, achieving most of the performance gains with moderate overhead, while \textit{PPAD-4r achieves the best results overall}.

\subsubsection{Visualization Results}
\label{sec:vis_case2}
\begin{wrapfigure}{r}{0.63\textwidth}
    \vspace{-0.35cm}
    \centering
    \includegraphics[width=0.63\textwidth]{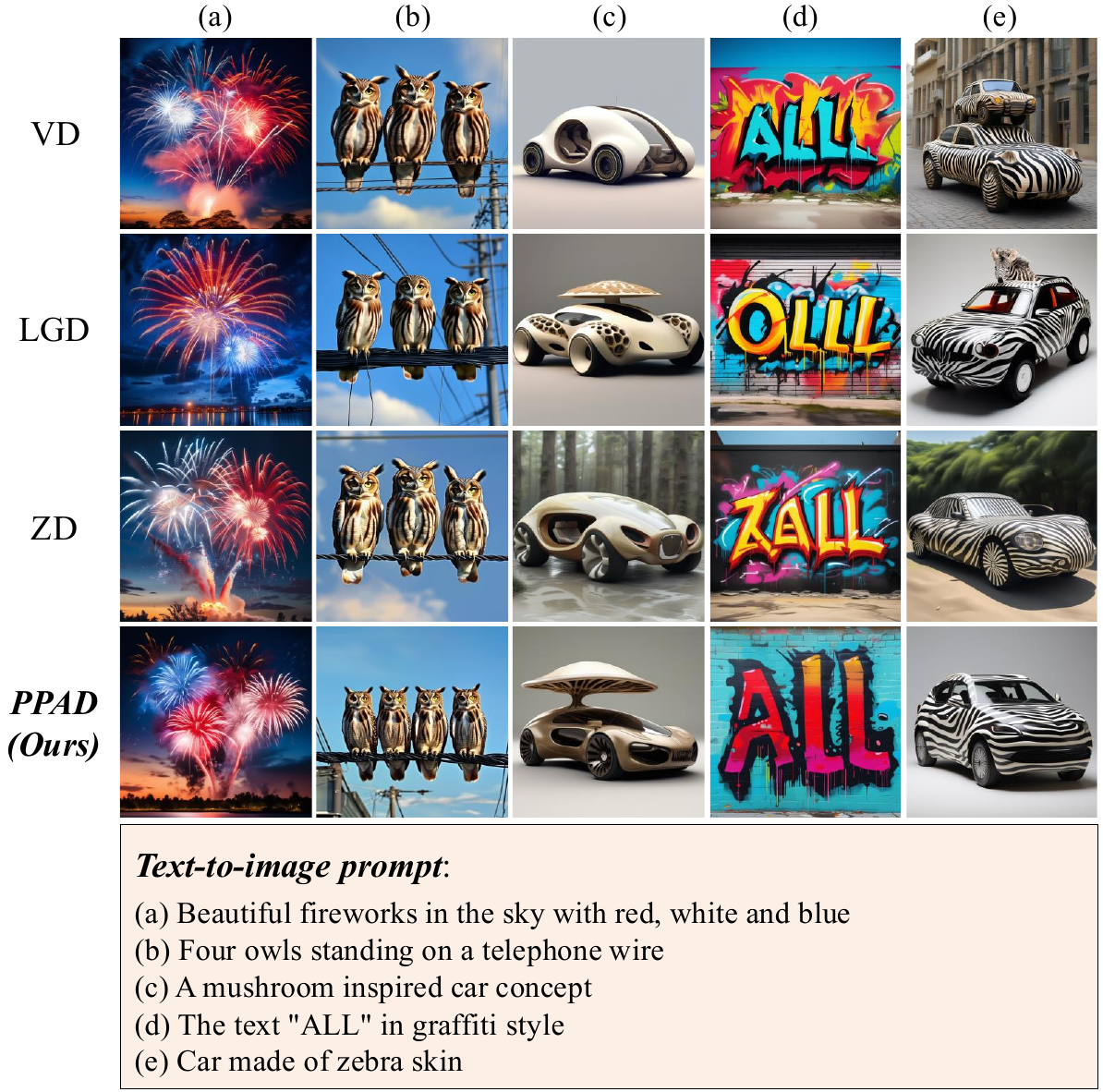}
    \vspace{-0.3cm}
    \caption{Comparison of generated images.}
    \label{fig:generation_case2}
    \vspace{-0.2cm}
\end{wrapfigure}

\noindent \textbf{Generation Comparisons.}
Figure~\ref{fig:generation_case2} illustrates a qualitative comparison between PPAD and three representative baselines across a diverse set of prompts, spanning semantics involving \textit{Color-Shape Binding}, \textit{Counting}, \textit{Abstract Concepts}, \textit{Text Rendering}, and \textit{Fine-Grained Attributes}.
\uline{(a) Color-Shape Binding.} This prompt requires associating multiple color attributes with the correct object (fireworks). All baselines struggle with maintaining fidelity across red, white, and blue simultaneously, often omitting or blending colors. In contrast, PPAD clearly depicts all three hues in well-formed, vivid fireworks.
\uline{(b) Counting.} This task involves recognizing and rendering four owls. Baselines frequently output only three or generate deformed figures. PPAD consistently generates four distinct, upright owls, showcasing strong alignment with count-based semantics.
\uline{(c) Abstract Concepts.} This prompt describes a highly imaginative concept — a mushroom-inspired car. While other methods loosely render organic or automotive shapes, only PPAD captures the hybrid aesthetic by integrating mushroom-like curves into the car’s structure.
\uline{(d) Text Rendering.} Text generation remains a longstanding challenge for diffusion models. Baselines produce misspelled or unreadable results (e.g., “OLLL”, “ZALL”), whereas PPAD alone generates a crisp, legible “ALL” in a graffiti style, demonstrating superior symbolic alignment.
\uline{(e) Fine-Grained Attribute.} The final prompt specifies a car made of zebra skin. Other methods fail to preserve consistent zebra patterns or distort the car shape. PPAD achieves a coherent composition where both the texture and object category are correctly rendered.
\noindent In summary, PPAD’s integration of \textit{semantic feedback}, \textit{refined prompt} adjustments, and correction for \textit{omission highlights} results in more accurate, faithful, and semantically aligned generations compared to existing baselines.

\begin{figure}[!ht]
    \centering
    \includegraphics[width=\textwidth]{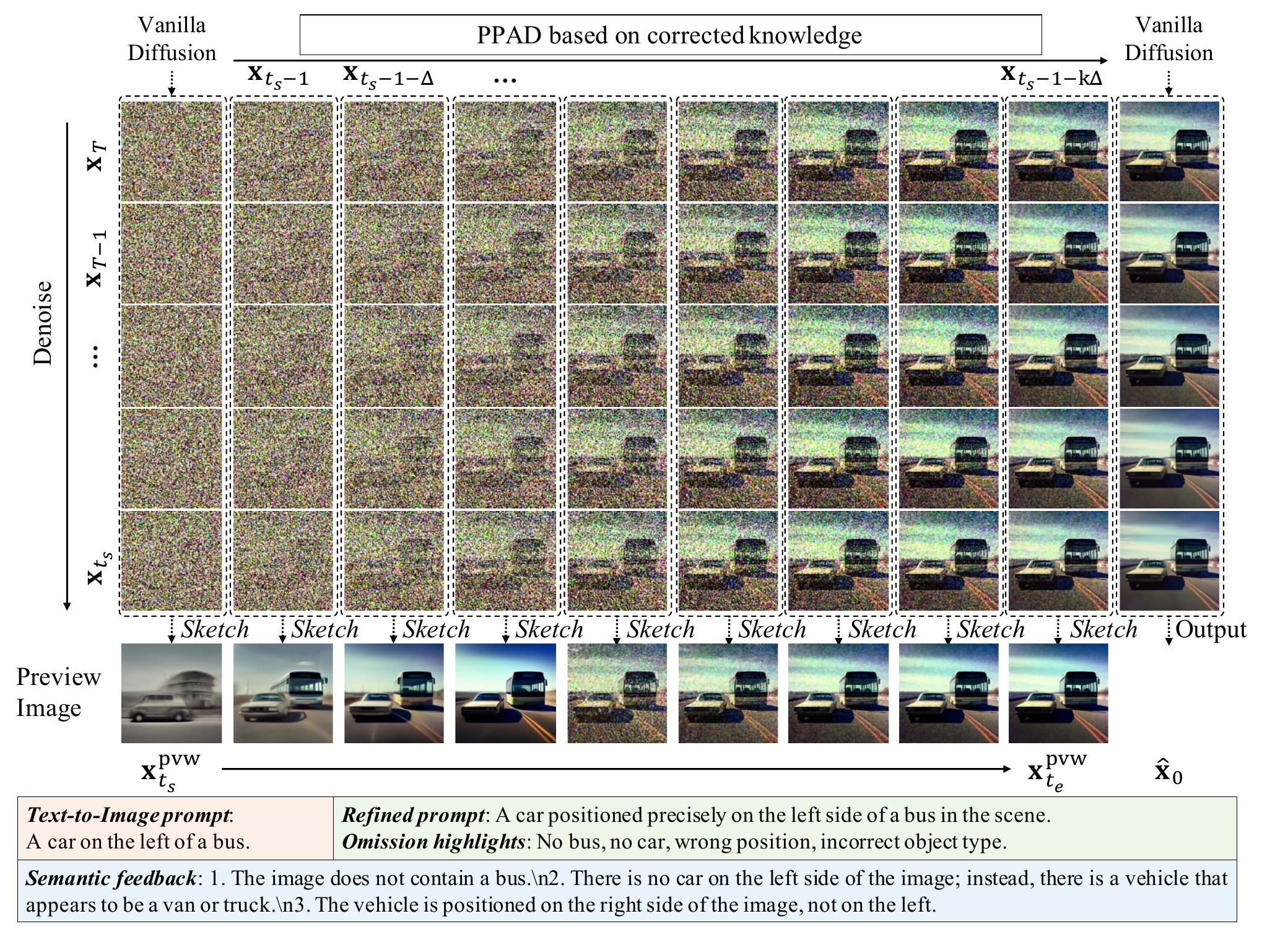}
    \vspace{-0.35cm}
    \caption{Denoising path with spatial omission and misplacement.}
    \label{fig:path_case2}
    \vspace{-0.3cm}
\end{figure}

\begin{figure}[!ht]
    \centering
    \includegraphics[width=\textwidth]{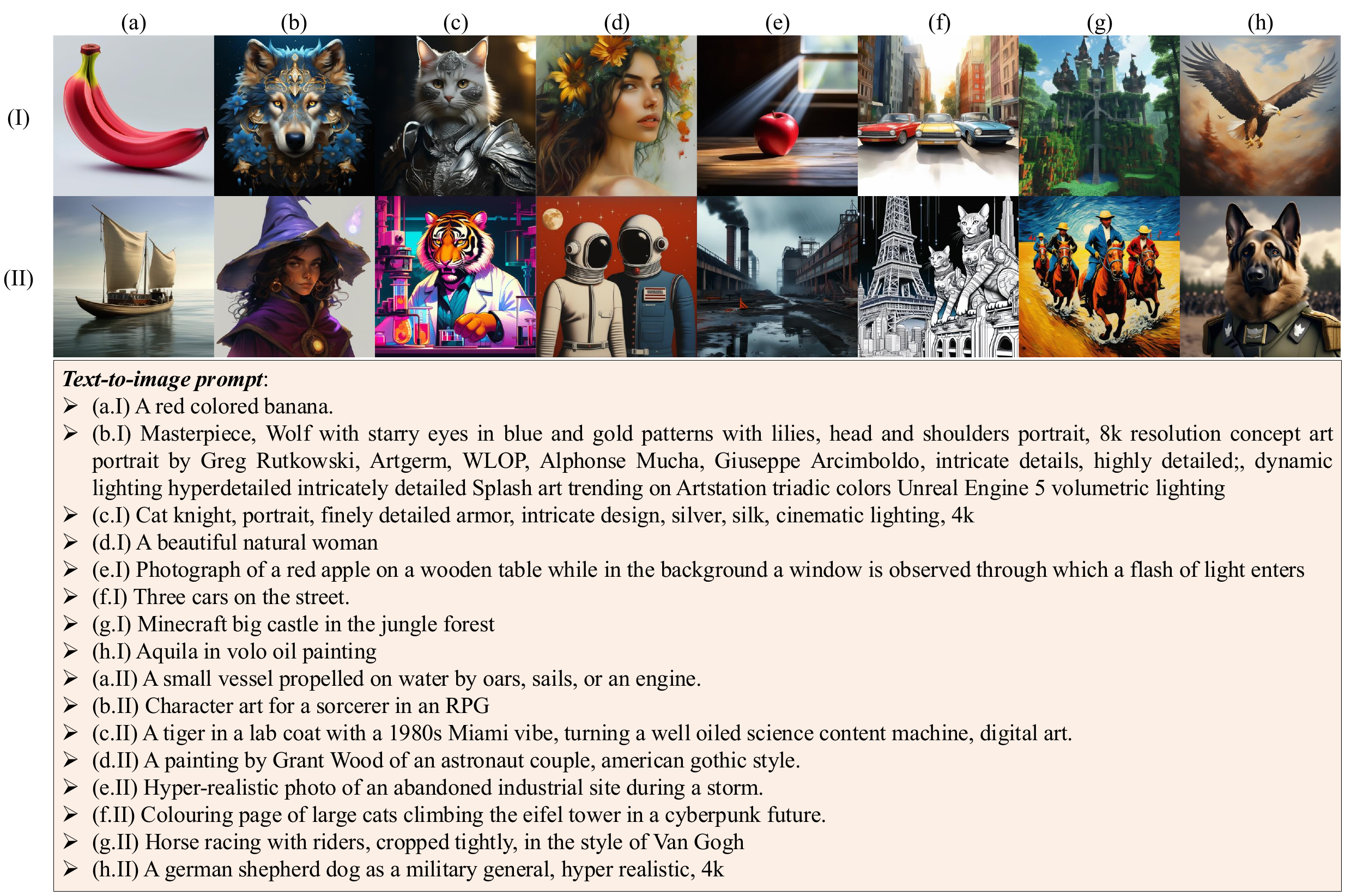}
    \vspace{-0.15cm}
    \caption{Generated samples conditioned on diverse prompts.}
    \label{fig:samples}
    \vspace{-0.3cm}
\end{figure}

\noindent \textbf{Denoising Path.} Figure~\ref{fig:path_case2} demonstrates a typical failure case involving spatial omission and misplacement, where a car should be positioned to the left of a bus. At step $\mathbf{x}{t_s}$, a preview image $\mathbf{x}^{\text{pvw}}{t_s}$ is generated and assessed by the MLLM, which detects three major issues: (1) the absence of a bus, (2) the absence of a car on the left, and (3) incorrect object positioning. These observations lead to a refined prompt that explicitly enforces the intended spatial layout, along with corresponding omission highlights.
From $\mathbf{x}{t_s}$ onward, the PPAD module re-guides the denoising path by incorporating the corrected prompt. During this interval, semantic alignment improves progressively—evident in the sketches between $\mathbf{x}{t_s}$ and $\mathbf{x}{t_e}$. The MLLM continues to provide semantic feedback, ensuring better alignment with the intended visual semantics. After reaching $\mathbf{x}{t_e}$, the final denoising steps follow vanilla diffusion to synthesize $\hat{\mathbf{x}}_0$. The final image shows clear alignment with the original prompt, validating the effectiveness of mid-process correction through semantic feedback.

\noindent \textbf{Multi-style Prompt Rendering.}
Figure~\ref{fig:samples} showcases a broad range of text-to-image samples generated by PPAD, illustrating its capability to generalize across diverse semantic and stylistic prompts. These prompts span artistic portraits, photorealistic renderings, imaginative scenes, and structured object layouts. Below, we categorize and analyze representative cases by row and column index:
\uline{(a.I) Basic Object Rendering.} The prompt involves a straightforward description: a red colored banana. This tests low-level visual alignment. PPAD correctly renders both the color and shape with minimal hallucination, demonstrating high visual grounding for simple entities.
\uline{(b.I, c.I, d.I)} \textit{Stylized Portraits.} These prompts involve intricate artistic descriptions or character concepts. (b.I) features a heavily stylized wolf in a fantasy aesthetic, while (c.I) and (d.I) involve a cat knight and a naturalistic woman, respectively. PPAD preserves key fine-grained details such as armor texture, facial features, and lighting, indicating its strength in rendering diverse character aesthetics.
\uline{(e.I, f.I)} \textit{Photorealism and Urban Scenes.} The red apple on the wooden table (e.I) demonstrates PPAD’s capacity for realism, successfully depicting natural lighting and depth-of-field. Meanwhile, (f.I) generates a plausible cityscape with three cars, showing PPAD’s ability to handle scene-level object compositionality.
\uline{(g.I, h.I)} \textit{Imaginative and Classical Styles.} The Minecraft-inspired jungle castle (g.I) and classical-style eagle oil painting (h.I) showcase PPAD’s adaptability to both voxel-art aesthetics and traditional painting textures, highlighting its stylistic flexibility.
\uline{(a.II, b.II, c.II)} \textit{Conceptual and Character Design.} These prompts range from a boat scene to RPG and digital-art styled characters. PPAD captures distinctive identities and scene composition with high semantic fidelity.
\uline{(d.II, e.II)} \textit{Narrative and Environmental Scenes.} The astronaut couple (d.II) is rendered in a style mimicking American Gothic, while (e.II) outputs a stormy industrial site with a high degree of atmospheric realism, demonstrating narrative-consistent generation.
\uline{(f.II, g.II, h.II)} \textit{Hybrid and Satirical Content.} These examples combine imaginative and humorous prompts (e.g., cyberpunk cats at Eiffel Tower, Van Gogh-styled horse race, German shepherd general). PPAD effectively maintains both thematic coherence and stylistic appropriateness, revealing its robustness in compositional generalization.
\noindent Overall, this figure evidences that PPAD is capable of consistent generation across a wide semantic spectrum, benefiting from its use of \textit{semantic feedback}, \textit{refined prompt} optimization, and recovery from \textit{omission highlights}. It enables faithful generation for both literal and abstract visual concepts, spanning multiple visual domains and artistic conventions.

\end{document}